\documentclass[lettersize,journal]{IEEEtran}
\usepackage{amsmath,amsfonts}
\usepackage{algorithm}
\usepackage{array}
\usepackage[table,dvipsnames]{xcolor}  
\usepackage[caption=false,font=normalsize,labelfont=sf,textfont=sf]{subfig}
\usepackage{textcomp}
\usepackage{stfloats}
\usepackage{url}
\usepackage{verbatim}
\usepackage{graphicx}
\usepackage{cite}
\usepackage{tcolorbox}

\usepackage{multirow}
\usepackage{makecell}
\usepackage{soul}
\usepackage[para,online,flushleft]{threeparttable}
\usepackage{booktabs}
\usepackage{ulem}
\usepackage{bm}
\usepackage[inkscapelatex=false]{svg}
\usepackage{nicematrix, tikz}
\usepackage{algpseudocode}
\usepackage{mdframed}

\newmdenv[
  backgroundcolor=yellow,
  linecolor=blue,
  linewidth=0.0pt,
  skipabove=0pt,
  skipbelow=0pt,
  innertopmargin=0pt,
  innerbottommargin=0pt,
  innerleftmargin=0pt,
  innerrightmargin=0pt
]{eqbox}

\begin{document}
	\title{SF-Loc: A Visual Mapping and Geo-Localization System based on Sparse Visual Structure  Frames
	}
	
	\author{Yuxuan Zhou, Xingxing Li*, Shengyu Li, Chunxi Xia, Xuanbin Wang, Shaoquan Feng 

		\thanks{This work was supported by the National Science Fund for
			Distinguished Young Scholars of China (42425401), the National
			Natural Science Foundation of China (423B2401), the
			National Key Research and Development Program of China (2023YFB3907100), and the China Postdoctoral Science Foundation-Hubei Joint 
			Support Program (2025T041HB). The numerical calculations
			in this paper have been done on the supercomputing system
			at the Supercomputing Center of Wuhan University.\textit{(Corresponding author: Xingxing Li.)}}
		\thanks{The authors are with School of Geodesy and Geomatics, Wuhan University, China  (e-mail: xingxingli@whu.edu.cn).}
	}
	
	\markboth{Journal of \LaTeX\ Class Files,~Vol.~14, No.~8, August~2021}%
	{Shell \MakeLowercase{\textit{et al.}}: A Sample Article Using IEEEtran.cls for IEEE Journals}
	
	\IEEEpubid{ }
	
	\maketitle
	\setlength{\abovecaptionskip}{2pt}
	\setlength{\belowcaptionskip}{0pt}
	\setlength{\textfloatsep}{12pt} 
	
	\begin{abstract}
	For high-level geo-spatial applications and intelligent robotics,  
	accurate global pose information is of crucial importance.  
	Map-aided localization is a generally applicable approach to overcome the limitations 
	of global navigation satellite system (GNSS) in challenging environments. 
	However, current solutions face challenges in terms of mapping efficiency, 
	storage burden and  re-localization performance. In this work, we present 
	SF-Loc, a lightweight visual mapping and map-aided localization system, 
	which is built upon the image-based map representation with dense but 
	compact depth, termed as visual structure frames. In the mapping phase, 
	multi-sensor dense bundle adjustment (MS-DBA) with global optimization 
	is applied to construct geo-referenced visual structure frames. 
	Besides, a co-visibility-based mechanism is developed to achieve incremental 
	mapping and keep the map sparsity. In the localization phase, coarse-to-fine 
	vision-based localization is performed, in which multi-frame information and 
	the map distribution are fully leveraged. To be specific, spatiotemporally 
	associated similarity (SAS) is proposed to overcome the place ambiguity, 
	and pairwise frame matching  is applied for  robust pose estimation. 
	Experimental results on the  real-world cross-temporal dataset verify the effectiveness 
	of the system. In complex urban road scenarios, the map size is down to 
	3 MB per kilometer and stable decimeter-level re-localization can be achieved.  
	 
	\end{abstract}
	
	
	\section{Introduction}
Global localization is a critical problem in robotics and 
autonomous driving applications \cite{ref1,ref2}. 
For outdoor scenarios, global navigation satellite 
system (GNSS) is a commonly used technology for 
achieving global drift-free localization\cite{ref_giv1}. 
However, precise GNSS generally relies on high-precision 
services and products, while low-cost GNSS can hardly 
provide stable and accurate localization solutions in 
complex environments\cite{ref_gnss}. Map matching is another 
method for  global localization. By associating observations  
from exteroceptive sensors (such as cameras)  mounted on intelligent 
devices with the prior maps, different levels of global localization can be achieved, depending on the map representation\cite{ref_map,ref_pr}.

To achieve the practicality of map-aided localization, 
both  mapping and  re-localization  need to be  handled. 
As to the mapping phase, the efficiency of map construction, 
the map accuracy, the storage burden and the information sufficiency for localization 
are some of the key factors. Current map forms, like 
vector map\cite{ref_bevlocator},  
3D point cloud\cite{ref_orb,ref_hloc}, 
occupancy grid\cite{ref_grid} and 
neural representations\cite{ref_locnerf,ref_splatloc}, 
still cannot fully satisfy all these requirements simultaneously, especially when 
considering vision-based low-cost schemes.

As to re-localization, the system functionality is highly 
coupled with the map data. High recall and high localization accuracy generally need  
more detailed and distinguishable information of the  
environment, which seems contradictory with the expectation 
of a lightweight map. 

It is noted that, the image itself can serve as an advantageous landmark for localization-oriented maps, which is rich in cues for fine-grained association,  widely accessible on ubiquitous devices, and  free from cross-modality information loss. However, the following issues should be considered:

\begin{enumerate}
	\item How to efficiently and accurately recover the absolute poses of images at a large scale?
	\item Since images are storage-intensive, to what extent can such image maps be compressed (or sparsified) while still meeting user localization requirements?
	\item What auxiliary information is needed for accurate 
	localization? 
	And  how to achieve high-recall localization in cross-temporal
	 or weak-recognizability scenarios?
\end{enumerate}

\begin{figure}[!t]
	\centering
	\includegraphics[width=8.4cm]{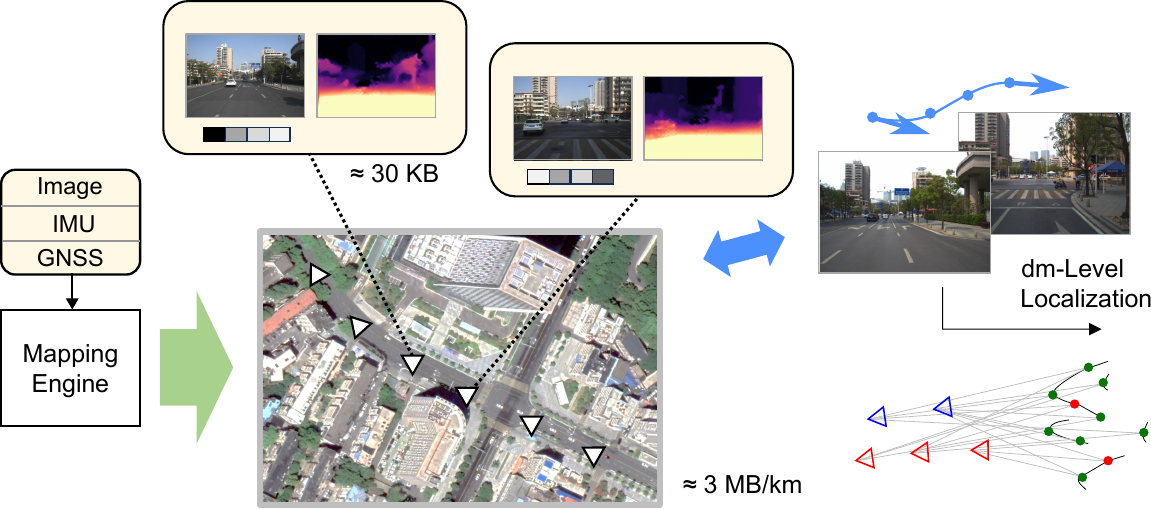}
	\caption{Illustration of the SF-Loc system. The system is built upon the map representation of visual structure frames, which contain compressed image, compact depth information and global descriptor. The map sparsity is intentionally maintained, which ensures lightweight storage ($\approx$ 3 MB/km) while keeps the ability of high-recall, decimeter level re-localization.}
	\label{fig_abstract}
\end{figure}

	In this work, we propose SF-Loc (as shown in Fig. \ref{fig_abstract}), a lightweight visual mapping and map-aided localization system based on sparse visual structure frames, aiming to comprehensively address the above issues. The visual structure frame is defined as a geo-tagged image frame with full-view depth, which itself supports standalone and flexible  re-localization.  Based on the concept, SF-Loc provides a holistically designed framework for efficient, accurate mapping and high-availability re-localization. The contributions of the paper are as follows:
	
		1) We provide a pipeline for large-scale mapping based on the visual-structure-frame map representation, which extracts accurate geometric information from global optimized multi-sensor dense bundle adjustment (MS-DBA) and supports the generation (incremental, multi-session, etc.) of practical localization-oriented maps.
		
	2) We provide a coarse-to-fine map-aided localization pipeline based on the structure frame map, which leverages the  spatiotemporally associated similarity (SAS) and pairwise frame matching for high-recall localization.
	
	3) Real-world, cross-temporal experiments are conducted to validate different phases of the system.
	
	4) The code is made open-source\footnote{ https://github.com/GREAT-WHU/SF-Loc} to benefit the community.

	\IEEEpubidadjcol
	
	\section{Related Work}
	
	\subsection{VSLAM with multi-sensor fusion}
	Visual simultaneous localization and mapping (SLAM), which has been extensively studied in the past decade, is considered an effective and low-cost solution for relative pose estimation and mapping. 
	By integrating IMU\cite{ref_vins,ref_okvis}, GNSS\cite{ref_giv1}, and other sensors\cite{ref_vow}, the practicality of VSLAM can be improved in terms of robustness, continuity, accuracy and geo-referencing functionality. On this basis, multi-session and multi-agent SLAM systems are developed\cite{ref_maplab2,ref_kimeram}. Learning-based VSLAM frameworks are also investigated in recent implementations, 
	covering neural/learnable representations including neural radiance fields (NeRF)\cite{ref_nicer} and 3D Gaussian splatting (3DGS)\cite{ref_gsslam}, data-driven association\cite{ref_dpvo,ref_droid,ref_ssfearture} and end-to-end pipelines\cite{ref_e2e_vo1}. However, to extract a practical reusable map from multi-sensor integrated VSLAM is still an open problem, especially when considering absolute accuracy 
	and cross-temporal functionality.


	\subsection{Visual place recognition}
	
	Visual place recognition (VPR) is a long-existing problem, which plays a crucial role in loop closure, multi-session mapping, re-localization and so on. 
	Early approaches use handcrafted models (like bag-of-words\cite{ref_bow} and VLAD\cite{ref_vlad}) to aggregate local feature descriptors, thus to evaluate the place similarity. To overcome the long-term change challenge, sequential information is utilized\cite{ref_seqslam}.
	
	In the last decade, deep neural network (DNN)-based methods, 
	like NetVLAD\cite{ref_netvlad} and its successors 
	with stronger  backbones\cite{ref_transVPR,ref_mixvpr},
	 have  greatly improved the practicality and accuracy of VPR.  
	 CosPlaces\cite{ref_cosplace} models place recognition as classification, which enables large-scale training and is further enhanced with better viewpoint robustness\cite{ref_eigenplaces}.
	AnyLoc\cite{ref_anyloc} utilizes the  visual foundation model to achieve ubiquitous 
	VPR for different scenarios. It is noted that, most existing 
	methods rarely exploit user-side temporal cues and database spatial structure, 
	which could serve as powerful information to improve the 
	stability of VPR.

	\subsection{Fine-grained visual re-localization}
	For high-accuracy visual re-localization (e.g. decimeter- or centimeter-level), it is generally needed to perform fine-grained association between user perception and the map elements.
	Many of these methods follow a coarse-to-fine pipeline. 
	In ORB-SLAM3\cite{ref_orb} and VINS-Mono\cite{ref_vins}, 
	re-localization is performed through bag-of-words-based 
	place recognition and point-level matching between feature 
	observations and the  point cloud map. Such pipeline is 
	further improved in HLoc\cite{ref_hloc}, where DNN-based 
	models are fully utilized. Besides, monocular depth prediction is exploited in \cite{ref_mapfree} to achieve map-free visual re-localization.
	
	Other approaches try to exploit higher-level information for map matching,
	 including instance-level object modeling\cite{ref_obj1,ref_obj4}, 
	 bird's-eye view (BEV) features\cite{ref_onet} and so on. 
	 The tradeoff of such semantic-based re-localization is the loss 
	 of detailed, sometimes subtle but distinguishable, 
	 visual cues, which might lead to higher ambiguity.

	 Moreover, relocalization based on neural/learnable representations (e.g., NeRF, 3DGS)  has recently attracted 
	 attention. However, existing work usually focuses on small 
	 scenes\cite{ref_locnerf,ref_splatloc} or image pose regression within the same sequence 
	 used for model construction\cite{ref_3dgsreloc}, 
		lacking practical studies on large-scale map-based  geo-localization.
	
	\section{System Overview}

	\begin{figure}[!t]
		\centering
		\includegraphics[width=7.4cm]{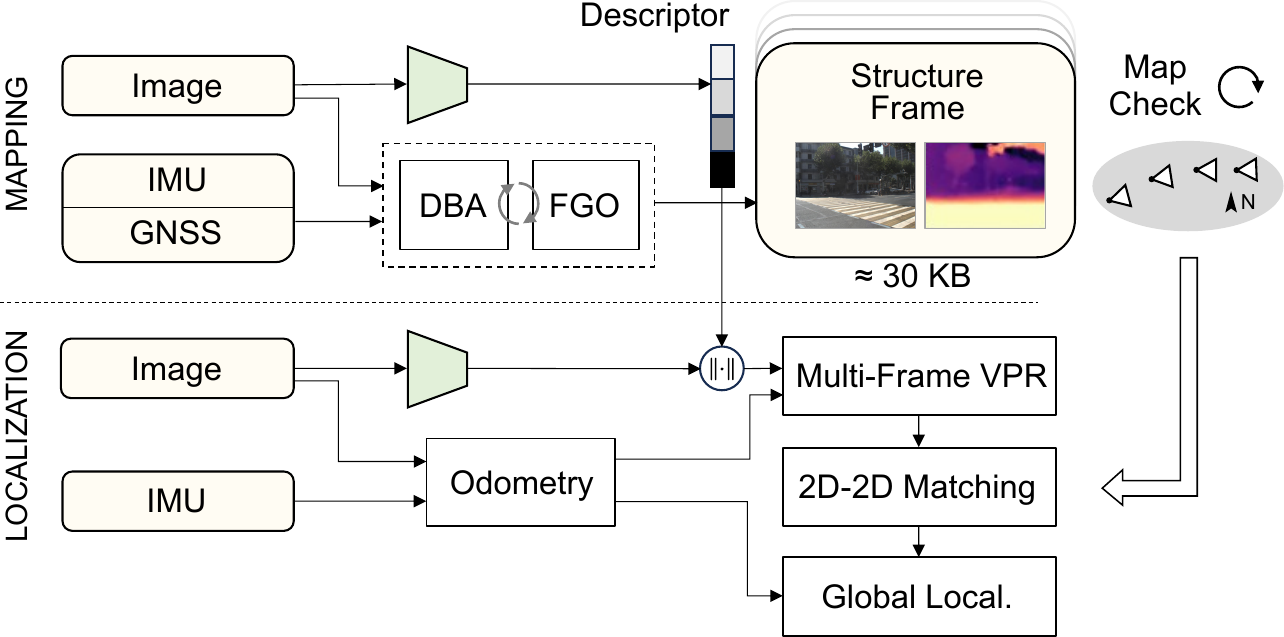}
		\caption{The overall pipeline of the system, which is divided into the mapping phase and the localization phase. }
		\label{fig_flow}
	\end{figure}

The proposed system is mainly divided into the mapping phase and the localization phase, as shown in Fig. \ref{fig_flow}. 

In the mapping phase, multi-sensor information is used to 
generate the feature-free map frames with recovered dense depth. 
Global optimization which incorporates local visual constraints is performed periodically at low frequency to achieve optimal pose estimation.
 On this basis, the local co-visibility is checked to keep the map sparsity and to achieve incremental mapping. Besides, a DNN-based VPR model is employed to compute the global descriptors of map frames.

In the localization phase, deep VPR and feature matching models are exploited to achieve coarse-to-fine localization. During this process, user-side multi-frame information and the map frame distribution are fully leveraged to improve the recall rate and the accuracy.

	
	\section{Mapping Phase}
	In this section, the implementation details of the mapping phase are presented.
	
	\subsection{Multi-Sensor Dense Bundle Adjustment}
	The foundation of the mapping process is to obtain accurate geometric information, i.e. the absolute poses and dense depths of the images. To achieve this, multi-sensor dense bundle adjustment (MS-DBA) with global optimization is employed.

		The concept of dense bundle adjustment (DBA) is proposed in\cite{ref_droid}. Specifically, sequential images are fed into a convolutional neural network (CNN) with gated recurrent units (GRUs) to compute  dense optical flows among co-visible image pairs. To explain this, consider the rigid flow produced by dense projection of one image pair (source frame $i$ projected to target frame $j$)
								\begin{equation}
			\mathbf{u}_{ij}=\Pi_c(\mathbf{T}_{ij} \circ \Pi_c^{-1} (\mathbf{u}_i,\bm{\lambda}_i)),\ \ 
			\mathbf{T}_{ij} = \mathbf{T}_j\circ \mathbf{T}_i^{-1}
			\label{DBA}
		\end{equation}
		where $\Pi_c$ is the camera projection model, $\mathbf{T}_{ij}$ is the 
		relative camera pose which lies in the 3D special Euclidean group SE(3), 
		``$\circ$'' is the transformation operation of SE(3),  
		$\mathbf{u}_i$ denotes the $n = (H/8) \times (W/8) $ grid-like 2D points in the 
		image frame which balance scene structure completeness and computational cost,
		 $\bm{\lambda}_i$ 
		is the inverse depth map. 
		
		The optical flow module takes the rigid flow and the CNN-encoded image feature map as  input and outputs the corrected optical flow (in the residual form) together with the  weight $\mathbf{w}_{ij}$, leading to the following linearized error equation
		\begin{equation}
 \left(\delta\mathbf{u}_{ij}\right)_{2n\times1} 
				=
\left[\setlength{\arraycolsep}{2.5pt}\begin{matrix}
	\mathbf{J}_i & \mathbf{J}_j & \mathbf{J}_{\bm{\lambda}_i}
\end{matrix}\right]
\left[\setlength{\arraycolsep}{2pt}\begin{matrix}
	\bm{\xi}_i^\top & \bm{\xi}_j^\top & \left({\delta\bm{\lambda}_i}\right)_{n\times1}^\top
\end{matrix}\right]^\top
		\end{equation}
		where  $\bm{\xi}_i$, $\bm{\xi}_j$ are  Lie algebras of the camera poses (corresponding to $\mathbf{T}_i$ and $\mathbf{T}_j$), $\mathbf{J}_i$, $\mathbf{J}_j$, $\mathbf{J}_{\bm{\lambda}_i}$ are Jacobians derived from (1).
		
		The residual-form dense optical flows are treated as re-projection errors, which are  used for bundle adjustment  to simultaneously estimate the poses and dense depths of the images. This process is recurrently performed and is end-to-end trainable, which shows good generalization in real-world scenes with synthetic training\cite{ref_droid}.
		
		In our early work\cite{ref_dbaf}, we demonstrate the possibility of 
		tightly integrating DBA with multi-sensor information. To be specific, 
		the Hessian form of Eq. (2) is derived   
		and stacked for all projections from frame $i$, leading to
			\begin{equation}
			\left[\begin{matrix}
				\mathbf{v}_{i} \\ \mathbf{z}_{i}
			\end{matrix}\right] = \left[\begin{matrix}
				\mathbf{B}_{i} & \mathbf{D}_{i} \\
				\mathbf{D}^\top_{i}  & \mathbf{C}_{i}
			\end{matrix}\right]
			\left[\begin{matrix}
				\bm{\xi}_{i,1,2,\dots,N}  \\ {\delta\bm{\lambda}_i}
			\end{matrix}\right]
		\end{equation}
	where $\mathbf{B}_i$, $\mathbf{C}_i$, $\mathbf{D}_i$ are  block matrices of the Hessian matrix,
	$\bm{\xi}_{i,1,2,\cdots,N}$ is the stacked Lie algebras of camera poses.
				
		The dense depth states are  further eliminated through Schur 
		complement to construct a compact constraint among the poses, following
		\begin{equation}
	(\underbrace{\mathbf{B}_i - 
	\mathbf{D}_i\mathbf{C}_i^{-1}\mathbf{D}_i^\top}_{\mathbf{H}_{c,i}})
	 \bm{\xi}_{i,1,2,\cdots,N} =  \underbrace{\mathbf{v}_i - \mathbf{D}_i 
	 \mathbf{C}_i^{-1} \mathbf{z}_i}_{\mathbf{v}_{c,i}}
	 \label{eq_hessian}
\end{equation}
in which  $\mathbf{C}_i$ is  a diagonal matrix.

	\begin{figure}[!t]
	\centering
	\includegraphics[width=7.4cm]{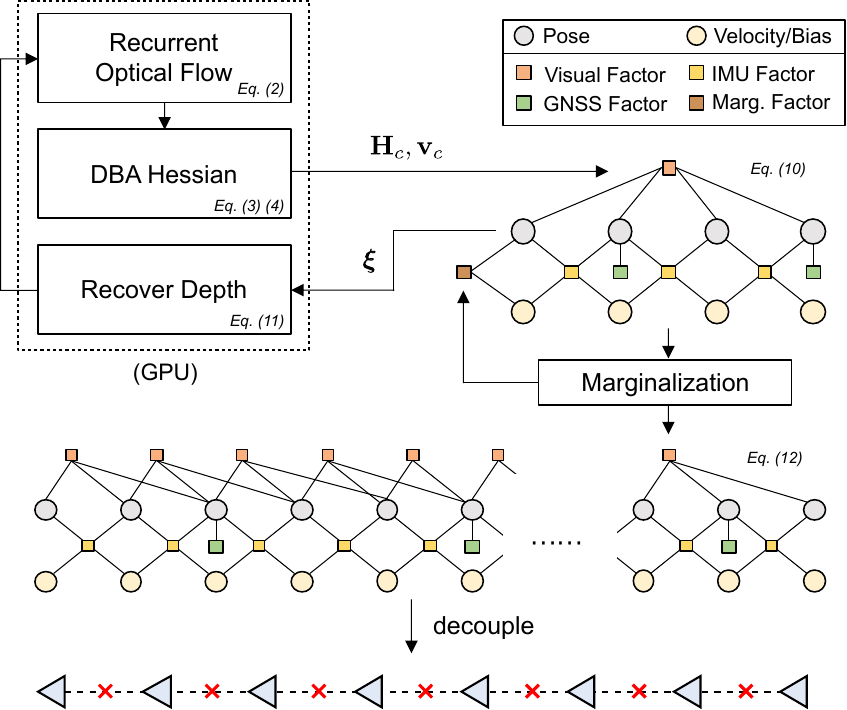}
	\caption{Illustration of the proposed multi-sensor DBA. 1) A sliding window factor graph is used for real-time state estimation and depth estimation, which is tightly integrated with the recurrent optical flow module. 2) The global factor graph collects the marginalized factors and provides low-frequency, long-time smoothed optimization results. 3) After mature, the frames in the global factor graph are decoupled to serve as standalone visual structure frames and inserted into the map. }
	\label{fig_dba}
\end{figure}

The Hessian-form pose constraint Eq. (\ref{eq_hessian}),
which embeds the DBA information of 
co-visible images, is fed to  a factor graph based on 
GTSAM\cite{ref_gtsam} for multi-sensor fusion, as shown in 
Fig. \ref{fig_dba}.  
A generic sliding-window graph design is employed, 
which allows for  flexible integration of different 
sensor information. The states $\mathbf{x}$ in the graph are as follows
\begin{equation}
	\mathbf{x} = \left\{\mathbf{x}_k \mid k\in\mathcal{K}\right\} ,\ \ 
				\mathbf{x}_k = 	
		\left(\begin{matrix}
			\mathbf{T}^w_{b_k} & \mathbf{v}^w_{b_k} & \mathbf{b}_{k}
		\end{matrix}
		\right)
					\end{equation}
		\begin{equation}
			\mathbf{T}^w_{b_k} = 	
			\left[\begin{matrix}
				\mathbf{R}^w_{b_k} & \mathbf{t}^w_{b_k} \\
				0& 1 
			\end{matrix}\right],\ \ \mathbf{b}_k = \left(\begin{matrix} \mathbf{b}_{a,k} & \mathbf{b}_{g,k} \end{matrix}\right)
		\end{equation}
		where $\mathbf{x}_k$ is the state at time $k$, $\mathcal{K}$ represents the set of timestamps within the sliding window, $\mathbf{T}^w_{b_k}$ is the body (IMU) poses in the world frame, $\mathbf{v}^w_{b_k}$ is the velocity, $\mathbf{b}_{a,k}$ and $\mathbf{b}_{g,k}$ are accelerometer/gyroscope biases. 
		
		The following factors are considered in this work:
		
		1) Visual factor:
			\begin{equation}
				\mathbf{E}_c (\mathbf{x}) =\frac{1}{2} l_c\!\left(\mathbf{x}\right)^\top\mathbf{H}_{c}\,l_c\!\left(\mathbf{x}\right) - l_c\!\left(\mathbf{x}\right)^\top \mathbf{v}_{c}
		\end{equation}
	  where $\mathbf{H}_{c}$, $\mathbf{v}_{c}$ are the 
	  Hessian matrix and vector of DBA 
	  (stacked from $\mathbf{H}_{c,i}$, $\mathbf{v}_{c,i}$ 
	  in Eq. (\ref{eq_hessian})) with all the depth states  
	  eliminated, $l_c(\cdot)$ is the linear container that 
	  transform the current states $\mathbf{x}$ to the 
	  Lie algebras of camera poses in Eq. (\ref{eq_hessian}) 
	  at specific linearization point.  
		
		2) IMU factor:
	\begin{align}
	\begin{split}
		&\mathbf{r}_{b}\left( \mathbf{x}_k,\mathbf{x}_{k+1}\right)= \\
		&\left[\begin{matrix}
			{\mathbf{R}^w_{b_k}}^{\top} \left( \mathbf{p}^w_{b_{k+1}}\! -\! \mathbf{p}^w_{b_k}\!+\! \frac{1}{2}\mathbf{g}^w\Delta{t^2_k}\! -\! \mathbf{v}^w_{b_k} \Delta t_k\!\right) -\! \Delta\tilde{\mathbf{p}}^{b_k}_{b_{k+1}}  \\
			{\mathbf{R}^w_{b_k}}^{\top} \left( \mathbf{v}^w_{b_{k+1}} + \mathbf{g}^w \Delta t_k - \mathbf{v}^w_{b_k}\right) - \Delta\tilde{\mathbf{v}} ^{b_k}_{b_{k+1}}  \\
			\text{Log}\left(\left({\mathbf{R}^w_{b_k}}\right)^{-1}  \mathbf{R}^w_{b_{k+1}}  {\left(\Delta\tilde{\mathbf{R}} ^{b_k}_{b_{k+1}}\right)^{-1}} \right) \\
			\mathbf{b}_{a,k+1} - \mathbf{b}_{a,k} \\
			\mathbf{b}_{g,k+1} - \mathbf{b}_{g,k}
		\end{matrix}\right]
	\end{split}
\end{align}
where $\Delta\tilde{\mathbf{p}}^{b_k}_{b_{k+1}}$, $\Delta\tilde{\mathbf{v}}^{b_k}_{b_{k+1}}$, $\Delta\tilde{\mathbf{R}}^{b_k}_{b_{k+1}}$ are the IMU preintegration terms\cite{ref_imu}, $\mathbf{g}^w$ is the gravity vector, $\Delta t_k$ is the time interval.
		
		3) GNSS factor:
	\begin{equation}
	\mathbf{r}_g (\mathbf{x}_k) =\mathbf{T}^n_w\circ \mathbf{T}^w_{b_k}\circ\mathbf{t}^b_g - \tilde{\mathbf{p}}^n_g
\end{equation}
where $\tilde{\mathbf{p}}^n_g$ is the measured position 
of  GNSS antenna in the navigation frame,
 $\mathbf{t}^b_g$ is the lever-arm, 
 $\mathbf{T}^n_w$ is a fixed world-to-navigation 
 transformation obtained from initial alignment.
		
		Probabilistic marginalization\cite{ref_okvis} is applicable to the above factor graph 
		based on Schur complement to ensure real-time performance. 
		Thus, the overall cost function of the sliding-window factor graph is defined as 
		follows
	\begin{align}
		\begin{split}
	\min\ \  \mathbf{E}_m (\mathbf{x}) + \mathbf{E}_c (\mathbf{x})& \\ + \sum_{k\in \mathcal{K}}\left\Vert \mathbf{r}_b(\mathbf{x}_k,\mathbf{x}_{k+1})\right\Vert^2_{\bm{\Omega}_b}
&	+ \sum_{k\in \mathcal{G}}
	\rho_C\left(
	\left\Vert
	 \mathbf{r}_g(\mathbf{x}_k)
	 \right\Vert^2_{\bm{\Omega}_g}
	 \right)
	 		\end{split}
\end{align}
where $\mathbf{E}_m(\mathbf{x})$ is the marginalization term in the quadric cost function form, $\mathcal{G}$ is the set of GNSS measurement epochs in the sliding window, $\bm{\Omega}_b$ and $\bm{\Omega}_g$ are uncertainties of the factors, $\rho_C(\cdot)$ is the Cauchy loss function.
		
		After the factor graph optimization, the corrected poses are substituted back to Eq. (3) to recover the dense depths
		\begin{equation}
			\delta \bm{\lambda}_i = \mathbf{C}_i^{-1} (\mathbf{z}_i-\mathbf{D}_i^\top\bm{\xi}_{i,1,2,\dots,N})
		\end{equation}
	then the estimated poses and depths are fed back into 
	the optical flow module to refine the optical flows $\delta\mathbf{u}_{ij}$
	and adjust the weights $\mathbf{w}_{ij}$. The overall process simultaneously adjust the pose estimation, the dense data association and the depths, trying to achieve consistency of the above information.
		
		For mapping tasks, an issue that should be considered is the obtainment of global optimal pose estimation.
		In realistic scenarios, both the observability 
		degradation of monocular visual-inertial odometry (VIO) and  intermittent 
		GNSS outages would lead to the decline of pose 
		accuracy, which can hardly be 
		overcome by a sliding-window estimator.  
		To handle this,  a global factor graph is 
		maintained in our system, which collects 
		the marginalized factors from the sliding-window 
		factor graph for further global optimization. 
		The objective function of the global factor graph is written as follows
			\begin{align}
				\begin{split}
			\min\ \ \sum_{k\in \mathcal{K}_\text{all}}\!\! \mathbf{E}_{c,k} (\mathbf{x}_{c,k})\!  +\!\! \sum_{k\in \mathcal{K}_\text{all}}\!\! \left\Vert \mathbf{r}_b(\mathbf{x}_k,\mathbf{x}_{k+1})\right\Vert ^2_{\bm{\Omega}_b}
		\!\! \\	+\!\! \sum_{k\in \mathcal{G}_\text{all}}\!\! \rho_C\left(\left\Vert \mathbf{r}_g(\mathbf{x}_k)\right\Vert ^2_{\bm{\Omega}_g}\right)
						\end{split}
		\end{align}
		where $\mathcal{K}_\text{all}$ is the set of timestamps in the global  
		graph, $\mathcal{G}_\text{all}$ is the set of 
		all GNSS measurement epochs, $\mathbf{E}_{c,k} (\mathbf{x}_{c,k})$ is 
		the marginalized DBA information taking  frame $k$ as the source frame (corresponding to Eq. (4)), $\mathbf{x}_{c,k}$ is the related poses. Each $\mathbf{E}_{c,k}$ reflects the local visual constraint around  frame $k$.
		
		This global factor graph includes all the information related to the poses, velocities and IMU biases, while it can still be solved efficiently thanks to the locality of the problem. Compared to pose-only pose graph estimation (PGO)\cite{ref_vins_fusion}, this graph contains complete IMU information and Hessian-form visual information, which can better smooth low-observability periods with future information.

\begin{figure}[!t]
	\centering
	\includegraphics[width=7.6cm]{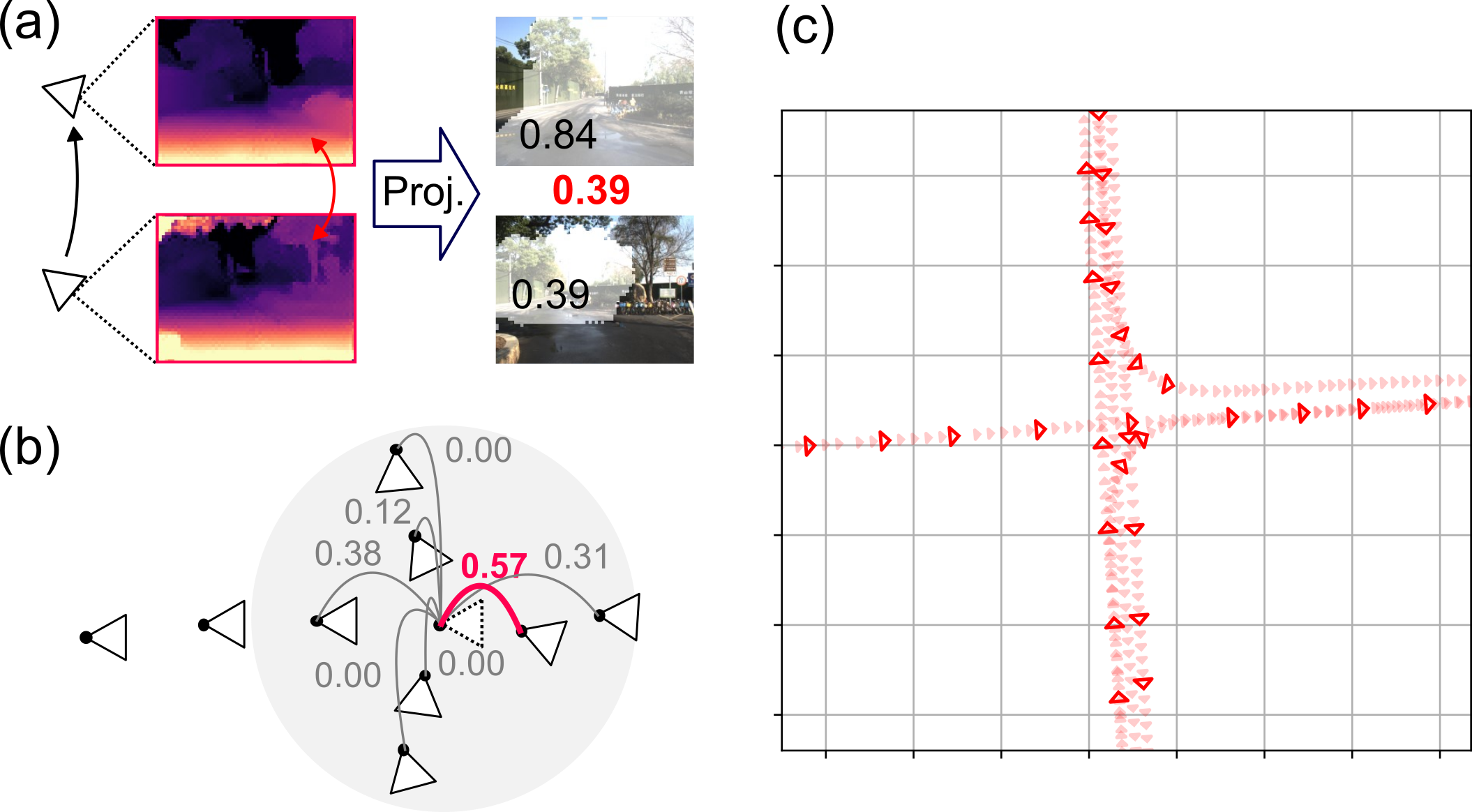}
	\caption{Illustration of the co-visibility checking. (a) Co-visibility for one image pair; (b) Local co-visibility checking, the dashed-line frame would be added/discarded based on its score comparison with the high-covisbility map frame; (c) An example of the sparsified structure frame map.  }
	\label{fig_cov}
\end{figure}

	\subsection{Map Construction and Maintenance}
	
	After global pose optimization, we extract the frame pose from the global graph and form a visual structure frame by combining the JPEG-compressed RGB image, low-resolution depth map, and global descriptor from the VPR model, which is subsequently inserted into the map.
	To make the map lightweight, it is important to keep 
	the sparsity of the visual structure 
	frames without significantly compromising 
	re-localization performance. Besides, considering the application scenario of multi-session mapping, 
	the map should support flexible extension and update. 
	
	Fortunately, this can be simply achieved based on the model-free, frame-based map representation of SF-Loc.
	To be specific, the local co-visibility is checked to preserve essential frames in the map. With the global poses and dense depths, the co-visibility is efficiently computed based on the bi-directional dense rigid flow that  falls within the field of view, following
	\begin{equation}
	\tau(\mathbf{T}_i,\mathbf{T}_j,\bm{\lambda}_i) = \frac{\text{count}(\mathbf{u}_{ij}.\text{inside}(0,0,H,W))}{H\cdot W} 
	\end{equation}
	\begin{equation}
	\text{covis}(\mathbf{T}_i,\! \mathbf{T}_j,\! \bm{\lambda}_i,\! \bm{\lambda}_j)\! =\! \min(	\tau(\mathbf{T}_i,\!\mathbf{T}_j,\!\bm{\lambda}_i),\!	\tau(\mathbf{T}_j,\!\mathbf{T}_i,\!\bm{\lambda}_j))
	\end{equation}
	where $\mathbf{u}_{ij}$ is the computed rigid flow in Eq. (1), $H$, $W$ are the height and width of the image. The illustration of the co-visibility is presented in Fig. \ref{fig_cov}.

	The co-visibility directly reflects the overlapping of different frames. With the intention of avoiding redundant map frames, the inter-frame co-visibility is controlled under a threshold $\Xi$. Considering adding/removing/updating frames, the following mechanism is employed:
	
	1)	When the mapping vehicle passes through a place and generates a new map frame, existing map frames within a certain range nearby are retrieved.
	
	2)	The bidirectional co-visibility is checked between the new map frame and other frames.
	
	3)	If the bidirectional co-visibility between the new map frame and all other frames is below $\Xi$, add the new map frame to the map.
	
	4)	If there exist map frames with bidirectional co-visibility above $\Xi$ with the new map frame, and the \textbf{score} of the new map frame is higher than that of the existing frames, delete the existing frames and add the new map frame to the map, which actually update the map locally; otherwise, discard the new map frame. 
		
		Here, the \textbf{score} can refer to metrics such as freshness, uncertainty or the proportion of dynamic objects.
	
	It is noted that the map update mechanism is not designed 
	to achieve the most complete scene coverage but  to provide relatively large coverage with low redundancy. Under different considerations, different score metrics and update policies can be employed. The co-visibility threshold $\Xi$ directly controls the map sparsity, which reflects the trade-off between data size and re-localization performance. We select $\Xi=0.4$ as the typical value, which indicates that for one query along the smooth trajectory between two nearby map frames, an optimistic $(1.0+0.4)/2 = 0.7$ overlapping  can be expected.

	When adding/removing/updating frames in the map, it is not necessary to perform data association across nearby frames or reconstruct the 3D model, as every visual structure frame works fully independently. This makes the mapping phase highly flexible, and can be easily applied to multi-session mapping or incremental map updating.

	\section{Map-Aided Localization Phase}
		In this section, the implementation details of the 
		map-aided localization are presented, which 
		follows the coarse-to-fine process like \cite{ref_hloc}. 
		Based on state-of-the-art place recognition and 
		feature matching models, specially designed methods 
		which consider spatiotemporal association are proposed to align with the map representation and to maximize the performance in realistic scenarios.
	
	\subsection{Multi-Frame Place Recognition}
	We use lightweight CNN-based methods (e.g., \cite{ref_netvlad,ref_mixvpr,ref_cosplace,ref_eigenplaces}) to compute global descriptors and to evaluate the similarity between the query image and map frames. On this basis, we develop a model-driven technique to integrate sequential queries and map frame distribution to improve the robustness and stability of the coarse localization.
	
	The necessity of multi-frame place recognition lies in the existence of repetitive and featureless scenes in the real world, which leads to inevitable spatial ambiguity. To handle this, it is natural to take into account the environmental cues within a larger range to resolve the ambiguity. Notice that we call this process ``multi-frame place recognition'' rather than ``sequential place recognition'', as the map frames are generally not ``sequential'' but are ``randomly'' distributed in the geographic space. Especially for multi-session mapping or at intersections, the spatially nearby map frames may come from multiple non-consecutive sequences.
	
	To avoid the ambiguity of single-shot VPR, we propose a metric called 
	spatiotemporally associated similarity (SAS) that combines the 
	information of appearance similarity, map distribution and 
	user-side relative poses. To be specific, we set particles around every landmark frames in the map, with every particle to be a 2D pose $(x,y,\theta)$ that represents a candidate of the query pose.  With the help of the user-side odometry, we compute the virtual  trajectory of every particle, following
	\begin{equation}
		\left(\mathbf{T}^{i}\right)_{4\times4\times L} = \mathbf{T}^w_{p_i} \left\{\mathbf{T}^{c_k}_{c_{k+1-j}}\right\}_{j=1,2,\cdots,L}, {i \in \mathcal{P}}
	\end{equation}
	where $\mathbf{T}^w_{p_i}$ is the particle pose, 
	$\left(\mathbf{T}^{i}\right)_{4\times4\times L}$ 
	is the virtual trajectory, $\mathcal{P}$ is the set of particle candidates, $L$ is the window size used for multi-frame VPR.
	
\begin{figure}[!t]
	\centering
	\includegraphics[width=7.5cm]{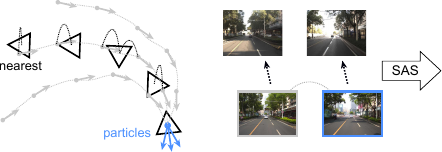}
	\caption{Illustration of the multi-frame place recognition. Particles that indicate the query pose are set around map frames. The user-side trajectory is used for nearest searching to associate with the map frames, then the SAS distance is evaluated, which combines the information of multiple image pairs. }
	\label{fig_mfvpr}
\end{figure}

	For every virtual trajectory, we efficiently search for the 
	spatially nearest map frame corresponding to every history 
	pose through KD-Tree, as shown in Fig. \ref{fig_mfvpr}. This leads to a sequence of L2 
	distances for every particle
	\begin{align}
		\left({\text{idx}}^i\right)_{L\times 1} &= \text{db}.\text{nearest}(\mathbf{T}^i),\ i \in \mathcal{P}\\
	\left(\bm{\sigma}^{i}\right)_{L\times1} &= \bm{\sigma}\left[  \text{idx}^i \right],\ i \in \mathcal{P}
	\end{align}
	where $\text{db}$ is the map database 
	organized by a spatial KD-Tree for fast searching, $\bm{\sigma}$ 
	is the $L \times M$ similarity matrix 
	which stores the L2 similarities between 
	the $L$ query frames and all $M$ frames 
	in the map, ``$[\cdot]$'' is the indexing operator. The similarity computation is performed every time a new query frame comes and stored in the memory,

	Then, we compute the SAS distance by averaging the L2 distances
	\begin{equation}
	\sigma^i_{\text{SAS}} = ||\bm{\sigma}^i||/\sqrt{L},\ i \in \mathcal{P}
	\end{equation}
	and the particle with minimum $\sigma^i_{\text{SAS}}$ is taken as the retrieved map frame.
	
	The rationale of SAS is based on the assumption that the historical 
	images of a correctly relocalized query should also exhibit high 
	similarity to their  spatially adjacent map frames, which helps 
	overcome the multipeaked characteristic of descriptor-based place 
	similarity evaluation function. The geographical distribution 
	of map frames, 
	along with user-side pose information, is implicitly incorporated 
	into the metric in nearest neighbor search. In this way, the 
	spatiotemporal association is naturally achieved,
	 without the need of an exact sequence-to-sequence matching. This technique can 
	 be easily adopted to common VPR models without extra layers or finetuning, 
	 transforming single-frame VPR similarities to multi-frame ones, which effectively broaden 
	 the perceived context for more discriminative recognition.
	 
	 In our implementation for outdoor environments, we simply set 3 particles 
	 around every map frame, corresponding to the 2D poses $(0,0,0^\circ)$, 
	 $(0,0,-30^\circ)$, $(0,0,30^\circ)$ relative to the map frame pose, which aligns with the 60$^\circ$ camera FOV and helps 
	 overcome the possible heading inconsistency of user/map frame poses.

	\begin{figure}[!t]
	\centering
	\includegraphics[width=8.0cm]{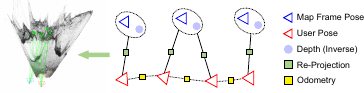}
	\caption{Illustration of the multi-frame fine localization. Pairwise query-to-map correspondences are computed and stored to construct a factor graph, in which the depth uncertainty and matching errors can be carefully handled.}
	\label{fig_fine_fgo}
\end{figure}

	\subsection{Fine Pose Estimation}
	After successful retrieval of the map frame corresponding to the query image, fine-grained re-localization can be performed. 
	
	Specifically, 2D-2D matching between the query and  map frames is
	performed based on either feature-based methods\cite{ref_lg} or detector-free methods\cite{ref_loftr}.
	With these correspondences and the full-view depth contained in the map frame, several techniques can be applied to estimate the pose of the query image, like perspective-n-point (PnP) or just aligning the two-view reconstruction result with the metric-scale depth.

	  Despite the simplicity, the above methods couldn't be directly used for 
	  multi-frame matching. In the application scenarios that our system concerns 
	  (cross-temporal, significant appearance variation, dynamic objects, etc.), robust feature matching is hard to promise. To provide accurate pose estimation under numerous outliers, we combine multi-frame correspondences and the odometry in a factor graph, as shown in Fig. \ref{fig_fine_fgo}. For better efficiency, we don't do exhaustive matching among all the related frames. Instead, we compute the correspondences between every query-to-database image pair from  VPR. Based on the re-projection errors, the cost function of the multi-frame pose estimation is written as follows
	\begin{align}
		\begin{split}
		\min\!\sum_{k\in \mathcal{K}}\!\!\left\Vert{\mathbf{T}^w_{c_{k+1}}}^{\!\!\!\!\!\!-1} \!\!\circ\!
		 \mathbf{T}^w_{c_k}
		  \!\!\circ\!
		   \tilde{\mathbf{T}}^{c_k}_{c_{k+1}}\! \right\Vert^2_{\bm{\Omega}_p}
		\!\!\!\! +\!\!\!
		 \sum_{k\in\mathcal{K}}\!\sum_{f\in\mathcal{F}_{k,i}}\!\!
		 \!\!\left\Vert \lambda^{m_i}_f 
		 \!\!-\!\!
		  \tilde{\lambda}^{m_i}_f\!\right\Vert^2_{\Omega_\lambda} \\
		  \!\!\! +\!\! \sum_{k\in \mathcal{K}}\!
		\sum_{f\in \mathcal{F}_{k,i}}\!
		\!\!\rho_C\!\!\left(\left\Vert \Pi_c\! \left(\!\mathbf{T}^{c_k}_{m_i}\!\!\circ\!{\Pi}_m^{-1}(\mathbf{u}^{m_i}_f,\lambda^{m_i}_f)\right)\!-\!\mathbf{u}^{c_k}_f\right\Vert^2_{\bm{\Omega}_c}
		\right)
		\end{split}
	\end{align}
	where $\tilde{\mathbf{T}}^{c_k}_{c_{k\!+\!1}}$ is the relative pose 
	estimation from the odometry, $\tilde{\lambda}^{m_i}_f$ is the sampled inverse 
	depth of point $f$ on the fixed map frame $m_i$, $\mathcal{F}_{k,i}$ is the matched feature point set between map frame $m_i$ and user-side frame $c_k$, $\mathbf{u}^{m_i}_f$ and $\mathbf{u}^{c_k}_f$ are the point coordinates respectively, ${\bm{\Omega}_p}$, ${\Omega_\lambda}$, ${\bm{\Omega}_c}$ are the uncertainties.

	\begin{figure}[!t]
	\centering
	\includegraphics[width=5.0cm]{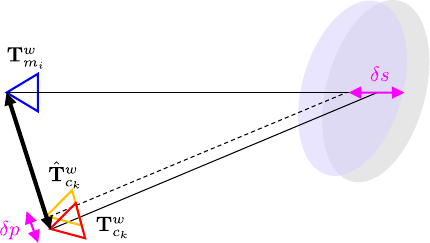}
	\caption{Illustration of how depth uncertainty of the map frame affects the user pose estimation accuracy. In the figure, $\delta s$ denotes the scale uncertainty of the depth map, $\delta p$ denotes the position error of user-side localization. An approximate $\delta p \propto \delta s \cdot  |\mathbf{t}^{m_i}_{c_k}|$ relationship can be found.  }
	\label{fig_depth}
\end{figure}

	 Another concern about the re-localization is whether the depth derived by 
	 monocular DBA sufficient for precise re-localization. An intuitive 
	 example is illustrated in Fig. \ref{fig_depth}. As is shown, 
	 the re-localization error is proportional to the product of 
	 the depth  error (simplified as scale error) and the recall distance. 
	 In vehicle-based multi-view triangulation, it is a moderate assumption 
	 to expect $<$10\% depth estimation uncertainty  for  objects within 50 m 
	 distance in most regions of the field of view (FOV). Given 10\% depth error 
	 of the map frame and $<$10 m recall distance, we can still expect a $<$1 m 
	 re-localization accuracy, which is adequate for our purposes. For the demand of robust re-localization, correct image 
	 retrieval and  feature matching matter more than the accuracy 
	 of structure information, allowing us to store the map 
	 frame depth in a highly compact way (i.e., $\times$8 downsampled, which naturally
	 aligns with the DBA process Eq. (\ref{DBA})).
	
	\section{Experiments}
	 	
	\begin{figure}[!t]
		\centering
		\includegraphics[width=8.0cm]{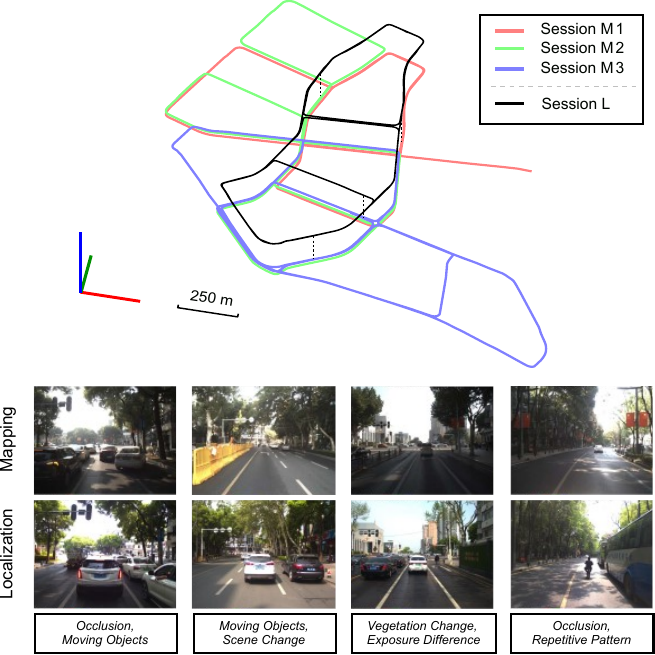}
		\caption{ Vehicle trajectories and example images of the experimental dataset. Three 1800-sec data sequences collected in 2022/10/23 (30 km mileage, 16 km one-way road coverage in total) are used for multi-session mapping, while the 1800-sec data sequence collected in 2023/04/12 (10 km mileage, 4.9 km one-way road coverage) is used for localization. Different trajectories are offset in the Z-direction to provide clearer visualization. }
		\label{fig_expr}
	\end{figure}

	In the experimental part,   
	different modules of the system, including multi-sensor mapping, 
	place recognition and fine re-localization are comprehensively tested. \footnote{All pretrained models and datasets are used under their respective licenses.}

	To fully evaluate the mapping and localization performance of the system in 
	 complex conditions, we use the self-made dataset collected in Wuhan City on 
	 2022/10/23 and 2023/04/12, 
	 featuring large scale, 
	 complete sensor coverage and reliable ground truth.
	  Specifically, three data sequences on 2022/10/23 are used to evaluate the performance of incremental mapping, which provides relatively complete coverage  of the region's road segments and exhibits multiple revisits. The data sequence on 2023/04/12 is used to evaluate the map-aided localization. Some challenging conditions are shown in Fig. \ref{fig_expr}, including exposure difference, occlusion, scene change and repetitive patterns.

The experimental vehicle is equipped with RGB cameras, an ADIS16470 IMU, a Septentrio AsteRx4 GNSS receiver and a tactical-grade IMU for reference use. The FOVs of the images used for mapping and localization are  different (60$^\circ$ vs. 57$^\circ$ horizontally). A nearby base station is used for differential GNSS (DGNSS) processing. The smoothed trajectory of post-processing DGNSS/INS integration based on high-end IMU is taken as  the ground truth, which is featured with centimeter-to-decimeter level position accuracy. 

 In the mapping phase, 5 Hz images (512 $\times$ 384) are used. In the localization phase, 1 Hz images (512 $\times$ 384)  are used.
  For multi-frame localization, except for the current frame, almost stationary frames are skipped. Considering the practical requirements of geo-localization applications, the focus is solely on horizontal positioning accuracy.

\subsection{Mapping Performance}

\begin{figure}[!t]
	\centering
	\includegraphics[width=8.0cm]{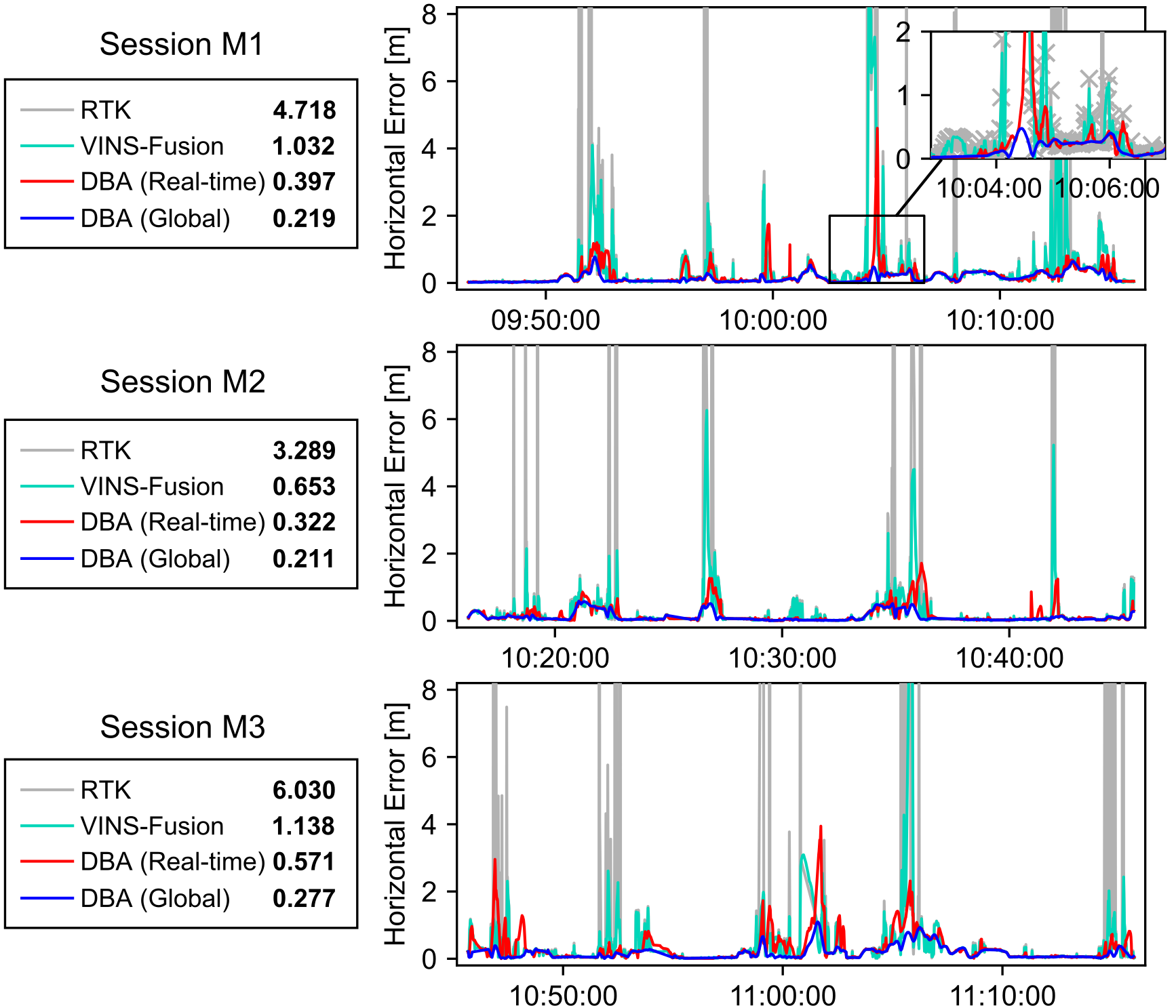}
	\caption{Pose estimation results of the mapping phase. Monocular schemes are used for vision-based methods. The bold numbers are root mean square errors (RMSE) of the horizontal position estimation.}
	\label{fig_mapping}
\end{figure}

\begin{figure}[!t]
	\centering
	\includegraphics[width=8.0cm]{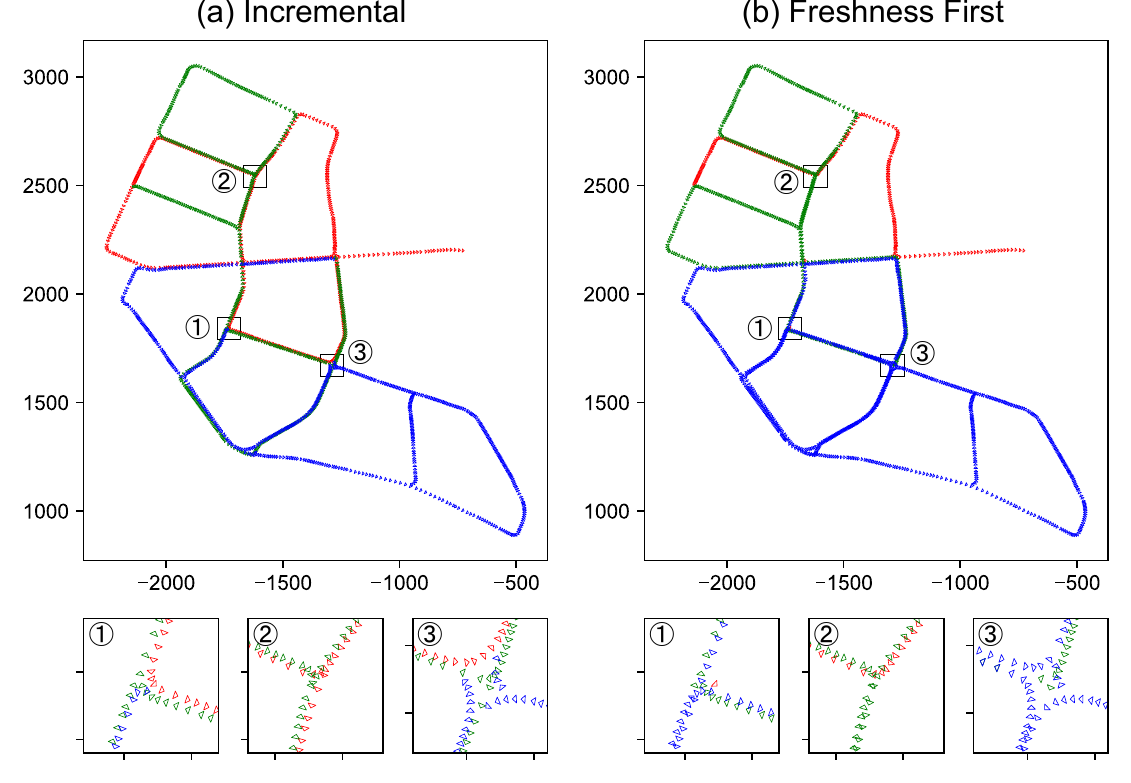}
	\caption{Multi-session mapping results (co-visbility threshold $\Xi=0.4$) with 
	different policies. (a) \textbf{Incremental:} In this mode, the frames of newer sessions have same scores with the old frames, thus the map would only be extended when new viewpoints are available. (b) \textbf{Freshness first:} In this mode, the frames of newer sessions have higher scores than the old frames, and the map would be updated when the vehicle passes through pre-built roads in newer sessions.}
	\label{fig_policy}
\end{figure}

We first focus on the effectiveness of the mapping phase. The data of 
different sessions are processed separately.

For large-scale mapping, obtaining accurate global poses is fundamental. Nevertheless, this is a challenging task considering the intermittent degradation of GNSS observations in the experimental scenario. In Fig. \ref{fig_mapping}, we show the pose estimation results of different approaches, including: 1) GNSS RTK, 2) VINS-Fusion (monocular VIO + RTK, final optimized trajectory) and 3) the proposed multi-sensor DBA scheme in real-time mode and global mode respectively. 

The results shown in Fig. \ref{fig_mapping} demonstrate that the DBA (global) scheme offers effective resilience to GNSS and VIO degradation, providing improved performance compared to VINS-Fusion and DBA (real-time) with RMSE reductions of 75.0\% and 45.2\% respectively. With the global optimization presented in Sect. IV-A, the position estimation error the DBA (global) scheme is mostly kept to decimeter-level throughout the sequence, which ensures the reference accuracy for user-side absolute localization.

Based on the global poses and dense depths, co-visibility-based multi-session mapping is performed. The distribution of generated map frames following different policies are shown in Fig.\ref{fig_policy}. The results demonstrate the effectiveness and flexibility of the proposed method in performing low-redundancy mapping under different criteria. In later discussions, we simply use the map generated in ``incremental'' mode for evaluation.

			\begin{table}[t]
				\centering
		\caption{Map storage of Different Mapping Schemes. For SF-Loc, Different 
		Co-Visibility Thresholds ($\Xi=0.4/0.3$) are 
		Considered. In Calculating the Distance, Non-Repeating One-Way Roads are Considered.
	}
	\label{table_map}
	\setlength\tabcolsep{4pt} 
	\begin{threeparttable}
	\begin{tabular}{lllll}
		\toprule
		\makecell[c]{Map Scheme} &
		\makecell[c]{Storage \\ (MB)} &
		\makecell[c]{Storage \\ per Km (MB)} &
		\makecell[c]{Frame \\ Number} &
		\makecell[c]{Average \\ Distance (m)} \\ 
		\midrule
		ORBSLAM3\ \ \ \ \ \ & \makecell[c]{6590.51}\ \textsuperscript{1} & \makecell[c]{403.53} & \makecell[c]{18768} & \makecell[c]{0.870} \\
		HLoc (SuperPoint)& \makecell[c]{32206.7}\ \textsuperscript{2}  &  \makecell[c]{1971.99} &  \makecell[c]{27026} &   \makecell[c]{0.607} \\
		\textbf{SF-Loc} ($\Xi$=0.4) & \makecell[c]{48.473}\ \textsuperscript{3} & \makecell[c]{2.967} & \makecell[c]{1727}& \makecell[c]{9.46} \\
		\textbf{SF-Loc} ($\Xi$=0.3) & \makecell[c]{33.007}\ \textsuperscript{4} & \makecell[c]{2.021} & \makecell[c]{1172}& \makecell[c]{13.94} \\
		\bottomrule
	\end{tabular}
	\begin{tablenotes}
		\scriptsize
		\textsuperscript{1} 1439  MB (Local Desc.) + 719.6 MB (Global Desc.) + 4431 MB (Model)\\
		\textsuperscript{2} 27261 MB (Local Desc.) + 106.7 MB (Global Desc.) + 4839 MB (Model)\\
		\textsuperscript{3} 37.37 MB (JPEG, 60) + 3.37 MB (Global Desc.) + 7.72 MB (Depth)\\
		\textsuperscript{4} 25.47 MB (JPEG, 60) + 2.29 MB (Global Desc.) + 5.24 MB (Depth)\\
	\end{tablenotes}
	\end{threeparttable}
\end{table}

 
 \begin{figure*}[!t]
 	\centering
 	\includegraphics[width=14.0cm]{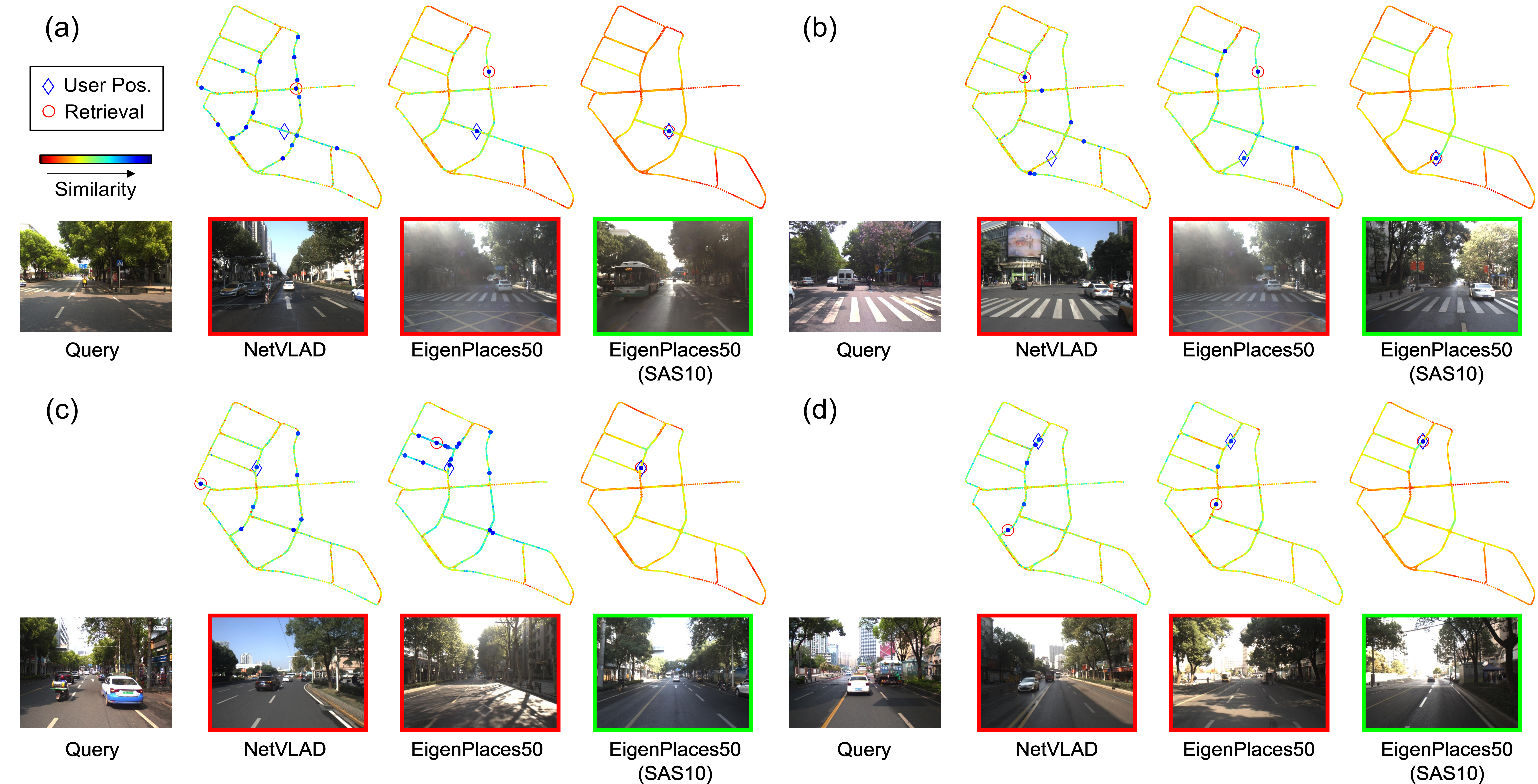}
 	\caption{Example cases of the coarse localization ($\Xi = 0.4$). The green borderline indicates correct retrieval, while the red borderline indicates wrong retrieval (large position error or heading error). High-similarity candidates (local minima and $\sigma <{\sigma}_{\text{min}} + 0.15( {\sigma}_{\text{max}} - {\sigma}_{\text{min}})$) are marked using  larger circles. }
 	\label{fig_coarse_case}
 \end{figure*}

Furthermore, the generated map data is analyzed. As baselines, 
we test ORBSLAM3 and HLoc for comparison, which are both 
commonly used pipelines for visual mapping and re-localization. 
For methods based on neural/learnable representations (e.g., NeRF
\cite{ref_locnerf}, 
3DGS\cite{ref_3dgsreloc,ref_splatloc}), 
as existing implementations lack support for such large-scale data (30 km length) and 
are not directly applicable to geo-localization tasks, they are not included in the evaluation.
For ORBSLAM3, we  use the monocular visual-inertial mode to process 
the sequence and save the map data in the Atlas format. The map storage 
is over 6 GB for the whole sequence, which is relatively heavy for 
practical applications. As for HLoc, we found it hard to achieve such 
large-scale SfM through the default COLMAP\cite{ref_colmap} pipeline. 
Instead, we directly use the ground-truth camera poses and perform feature extraction, matching and triangulation based on Superpoint\cite{ref_sp} and SuperGlue\cite{ref_sg}. The map storage is over 30 GB, which is 5 times larger than the ORBSLAM3 map. Such heavy map makes it hard to be applied to large-scale scenarios.

 As for SF-Loc, the map storage is much lighter. 
 As mentioned in Sect. IV-B, the co-visibility threshold $\Xi$ directly 
 controls the map sparsity. We test different co-visibility thresholds (0.3 or 0.4) 
 for map construction, which correspond to different 
 sparsities with necessary 
 consideration of scene coverage completeness. The final data size of the map is around 3 MB per kilometer 
 for $\Xi=0.4$ and around 2 MB per kilometer for $\Xi=0.3$, which is 1$\sim$3 
 orders of magnitude smaller than that of the aforementioned methods. It is noted that we use JPEG compression (quality = 60)
   for RGB storage and PNG compression (16 bit) for depth storage, which balances the
   data size and localization accuracy. 
 Later, we will discuss how high-precision localization can be achieved under such sparsified and compressed map information.

\renewcommand{\arraystretch}{0.9} 
\begin{table}[t]
	\centering
	\caption{Recall and Accuracy of the Coarse Re-Localization. 
	Maps with Different Sparsities (Co-Visbility Threshold $\Xi=0.4/0.3$) are Considered Respectively. }
	\label{table_coarse}
	\setlength\tabcolsep{4pt} 
	\begin{threeparttable}
		\begin{tabular}{clcccc}
			\toprule
			\makecell[c]{$\Xi$} &
			\makecell[c]{Method} &
			\makecell[c]{Recall\\@5 m} &
			\makecell[c]{Recall\\@10 m} & 
			\makecell[c]{Recall\\@20 m} & 
			\makecell[c]{RMSE\textsuperscript{1}\\(m)} \\ 
			\midrule
			
			\multirow{16}{*}{0.4} & ORB + DBoW2  & \makecell[c]{1.28\%} & \makecell[c]{2.01\%} & \makecell[c]{2.62\%} & \makecell[c]{49.17} \\
& NetVLAD & \makecell[c]{45.19\%} & \makecell[c]{59.53\%} & \makecell[c]{67.46\%} & \makecell[c]{15.01} \\
& AnyLoc & \makecell[c]{59.10\%} & \makecell[c]{80.39\%} & \makecell[c]{89.73\%} & \makecell[c]{12.99} \\
& MixVPR & \makecell[c]{65.89\%} & \makecell[c]{86.75\%} & \makecell[c]{95.60\%} & \makecell[c]{7.58} \\
& CosPlace18 & \makecell[c]{59.86\%} & \makecell[c]{84.41\%} & \makecell[c]{95.38\%} & \makecell[c]{8.22} \\
& CosPlace50 & \makecell[c]{61.54\%} & \makecell[c]{85.71\%} & \makecell[c]{95.55\%} & \makecell[c]{8.57} \\
& EigenPlaces18 & \makecell[c]{61.43\%} & \makecell[c]{84.57\%} & \makecell[c]{94.62\%} & \makecell[c]{8.35} \\
& EigenPlaces50 & \makecell[c]{63.88\%} & \makecell[c]{87.62\%} & \makecell[c]{97.12\%} & \makecell[c]{7.77} \\
			\cmidrule{2-6}
& EigenPlaces18 (5) & \makecell[c]{61.71\%} & \makecell[c]{85.44\%} & \makecell[c]{96.31\%} & \makecell[c]{9.28} \\
&  EigenPlaces50 (5) & \makecell[c]{64.31\%} & \makecell[c]{88.32\%} & \makecell[c]{98.21\%} & \makecell[c]{7.18} \\
&  EigenPlaces18 (10) & \makecell[c]{61.76\%} & \makecell[c]{85.61\%} & \makecell[c]{96.47\%} & \makecell[c]{8.75} \\
&  EigenPlaces50 (10) & \makecell[c]{64.48\%} & \makecell[c]{88.76\%} & \makecell[c]{98.75\%} & \makecell[c]{7.00} \\
			\cmidrule{2-6}
& EigenPlaces18 (SAS5) & \makecell[c]{74.69\%} & \makecell[c]{95.76\%} & \makecell[c]{99.57\%} & \makecell[c]{6.07} \\
& EigenPlaces50 (SAS5) & \makecell[c]{\textbf{74.74\%}} & \makecell[c]{96.36\%} & \makecell[c]{99.89\%} & \makecell[c]{4.78} \\
& EigenPlaces18 (SAS10) & \makecell[c]{74.47\%} & \makecell[c]{\textbf{97.01\%}} & \makecell[c]{\textbf{100.00\%}} & \makecell[c]{4.65} \\
& EigenPlaces50 (SAS10) & \makecell[c]{74.58\%} & \makecell[c]{96.80\%} & \makecell[c]{\textbf{100.00\%}} & \makecell[c]{\textbf{4.57}} \\
			
			\midrule
			
			\multirow{16}{*}{0.3} & ORB + DBoW2 & \makecell[c]{1.16\%} & \makecell[c]{1.77\%} & \makecell[c]{2.80\%} & \makecell[c]{47.78} \\
& NetVLAD & \makecell[c]{34.11\%} & \makecell[c]{51.44\%} & \makecell[c]{61.60\%} & \makecell[c]{17.45} \\
& AnyLoc & \makecell[c]{47.26\%} & \makecell[c]{76.10\%} & \makecell[c]{88.21\%} & \makecell[c]{13.82} \\
& MixVPR & \makecell[c]{52.69\%} & \makecell[c]{84.46\%} & \makecell[c]{95.27\%} & \makecell[c]{8.32} \\
& CosPlace18 & \makecell[c]{50.03\%} & \makecell[c]{82.24\%} & \makecell[c]{94.79\%} & \makecell[c]{8.88} \\
& CosPlace50 & \makecell[c]{50.30\%} & \makecell[c]{82.94\%} & \makecell[c]{95.17\%} & \makecell[c]{9.57} \\
& EigenPlaces18 & \makecell[c]{51.66\%} & \makecell[c]{83.76\%} & \makecell[c]{94.73\%} & \makecell[c]{8.26} \\
& EigenPlaces50 & \makecell[c]{53.50\%} & \makecell[c]{86.04\%} & \makecell[c]{96.80\%} & \makecell[c]{8.25} \\
			\cmidrule{2-6}
& EigenPlaces18 (5) & \makecell[c]{51.93\%} & \makecell[c]{84.41\%} & \makecell[c]{95.60\%} & \makecell[c]{9.49} \\
& EigenPlaces50 (5) & \makecell[c]{53.72\%} & \makecell[c]{86.91\%} & \makecell[c]{97.94\%} & \makecell[c]{7.97} \\
& EigenPlaces18 (10) & \makecell[c]{51.71\%} & \makecell[c]{84.36\%} & \makecell[c]{95.71\%} & \makecell[c]{8.44} \\
& EigenPlaces50 (10) & \makecell[c]{53.56\%} & \makecell[c]{86.64\%} & \makecell[c]{97.61\%} & \makecell[c]{7.96} \\
			\cmidrule{2-6}
& EigenPlaces18 (SAS5)& \makecell[c]{55.78\%} & \makecell[c]{91.80\%} & \makecell[c]{99.02\%} & \makecell[c]{6.70} \\
& EigenPlaces50 (SAS5) & \makecell[c]{56.60\%} & \makecell[c]{92.29\%} & \makecell[c]{99.57\%} & \makecell[c]{6.01} \\
& EigenPlaces18 (SAS10) & \makecell[c]{\textbf{56.93}\%} & \makecell[c]{\textbf{92.50}\%} & \makecell[c]{99.78\%} & \makecell[c]{5.97} \\
& EigenPlaces50 (SAS10) & \makecell[c]{56.49\%} & \makecell[c]{92.45\%} & \makecell[c]{\textbf{100.00\%}} & \makecell[c]{\textbf{5.80}} \\
			\bottomrule
		\end{tabular}
		\begin{tablenotes}
			\textsuperscript{1} For RMSE calculation, only $<$ 100 m cases are taken into account.
		\end{tablenotes}
	\end{threeparttable}
\end{table}

\renewcommand{\arraystretch}{0.9} 
\begin{table}[t]
	\centering
	\caption{Recall of the Coarse Re-Localization with Different Multi-Frame 
	Processing Techniques based on Different VPR Models. 
	The SF-Loc Map with $\Xi=0.4$ is Considered. }
	\label{table_aba_corase}
	\setlength\tabcolsep{4pt} 
	\begin{threeparttable}
		\begin{tabular}{clcccc}
			\toprule
			\makecell[c]{VPR\\Model} &
			\makecell[c]{Multi-\\Frame} &
			\makecell[c]{Recall\\@5 m (\%)} &
			\makecell[c]{Recall\\@10 m (\%)} & 
			\makecell[c]{Recall\\@20 m (\%)} \\ 
			\midrule
			
			\multirow{3}{*}{NetVLAD} & - & \makecell[c]{45.19} & \makecell[c]{59.53} & \makecell[c]{67.46} \\
			& 10 & \makecell[c]{49.05 (3.86↑)} & \makecell[c]{66.11 (6.58↑)} & \makecell[c]{76.75 (9.29↑)} \\
			& SAS10 & \makecell[c]{69.20 (\textbf{24.0}↑)} & \makecell[c]{91.31 (\textbf{31.8}↑)} & \makecell[c]{96.31 (\textbf{28.9}↑)} \\
			\midrule
			\multirow{3}{*}{AnyLoc}& - & \makecell[c]{59.10} & \makecell[c]{80.39} & \makecell[c]{89.73}  \\
			& 10& \makecell[c]{60.24 (1.14↑)} & \makecell[c]{82.07 (1.48↑)} & \makecell[c]{92.02 (2.29↑)}  \\
			& SAS10& \makecell[c]{71.92 (\textbf{12.8}↑)} & \makecell[c]{95.60 (\textbf{15.2}↑)} & \makecell[c]{99.51 (\textbf{9.78}↑)}  \\
			
			\midrule
			
			\multirow{3}{*}{MixVPR} & - &\makecell[c]{65.89} & \makecell[c]{86.75} & \makecell[c]{95.60}  \\
			& 10 & \makecell[c]{66.16 (0.27↑)} & \makecell[c]{87.29 (0.54↑)} & \makecell[c]{96.52 (0.92↑)}  \\
			& SAS10 & \makecell[c]{72.30 (\textbf{6.41}↑)} & \makecell[c]{96.36 (\textbf{9.61}↑)} & \makecell[c]{99.95 (\textbf{4.35}↑)}
			
 \\

			\bottomrule
		\end{tabular}
	\end{threeparttable}
\end{table}

\subsection{Localization Performance}
As to user-side localization, 
we check the coarse-to-fine 
localization performance of the proposed 
system. Considering realistic cases of 
localization query, we test single-frame and multi-frame (with temporal fusion) localization performance, while historical prior information (like Kalman filtering) is not considered.

\subsubsection{Coarse Localization}
The test of coarse localization is based on the map frames generated by SF-Loc. 
The coarse localization performance directly affects further fine-grained 
association, and a $<$20 m recall distance is generally needed for 
possible fine localization under the experimental scenario. We show the 
recall rates of different schemes in TABLE \ref{table_coarse}, including 
the classic ORB + DBoW2 implementation\cite{ref_orb} and DNN-based methods 
(NetVLAD\cite{ref_netvlad}, AnyLoc\cite{ref_anyloc}, MixVPR\cite{ref_mixvpr}, 
CosPlace\cite{ref_cosplace}, EigenPlaces\cite{ref_eigenplaces}). 
For CosPlace and EigenPlaces, different backbones 
(ResNet18 or ResNet50\cite{ref_resnet}) are considered.
Based on the state-of-the-art lightweight  model EigenPlaces, we 
employ the proposed SAS technique to aggregate the information of 
the current and last $K$ sequential queries, termed as ``EigenPlaces (SAS$K$)'', where different window lengths are considered.
 To verify the effectiveness of the proposed SAS metric, we also implement a naive multi-frame VPR model by clustering the top-$10$ candidates of the current and last $K$ sequential queries and choosing the candidate belonging to the largest cluster as the result, termed as ``EigenPlaces ($K$)''.

Taking the $\Xi=0.4$ map as example, several typical query cases are shown 
in Fig. \ref{fig_coarse_case}, which indicate 
the challenge of  re-localization, with (a) lighting condition variation, 
(b) seasonal vegetation change, 
(c) dynamic objects,
 and (d) high-ambiguity environmental 
patterns.  In such scenarios, the traditional handcrafted place recognition method 
(ORB + DBoW2) becomes almost useless, as the low-level intensities of the environment 
changes significantly for the cross-temporal sequences. 
For learning-based methods, the classic NetVLAD achieves 
59.53\% 10 m recall, while more modern methods 
(e.g., CosPlace, EigenPlaces and MixVPR) can 
generally achieve 85\% 10 m recall. Although 85\% 
recall seems good enough for VPR, 
it can't serve as a reliable information source 
for the localization task, especially for unmanned 
systems. By naively combining the VPR results of 
multiple frames, the 10 m/20 m recall can increase 
by around 1\%. In contrast, with the proposed 
SAS-based multi-frame VPR technique, the 10 m/20 m 
recall increases by approximately 10\%/5\%. For the 
SAS10 scheme, the 20 m recall achieves 100\%  for 
both ResNet50 and ResNet18 backbones. Compared to 
single-frame VPR methods, the introduction of SAS 
overcomes the challenges depicted in Fig. \ref{fig_coarse_case}
and greatly improves the reliability of the coarse 
localization, which is crucial for further fine 
localization. Essentially, the association of 
spatiotemporal information serves as a cross-verification 
mechanism, enhancing the confidence of  visual 
cues and leading to improved place recognition 
performance under significant appearance changes.

We also list the results based on a sparser map with co-visibility threshold $\Xi=0.3$. The recall is noticeably lower than the $\Xi=0.4$ case, while the EigenPlaces50 (SAS10) scheme can still achieve 92.45\%/100\% 10 m/20 m recall. 

As extra reference, we employ the SAS technique to  VPR models other than EigenPlaces. 
The results listed in TABLE \ref{table_fine} verify its broad applicability and superiority over simple clustering-based multi-frame VPR.

\begin{figure}[!t]
	\centering
	\includegraphics[width=8.0cm]{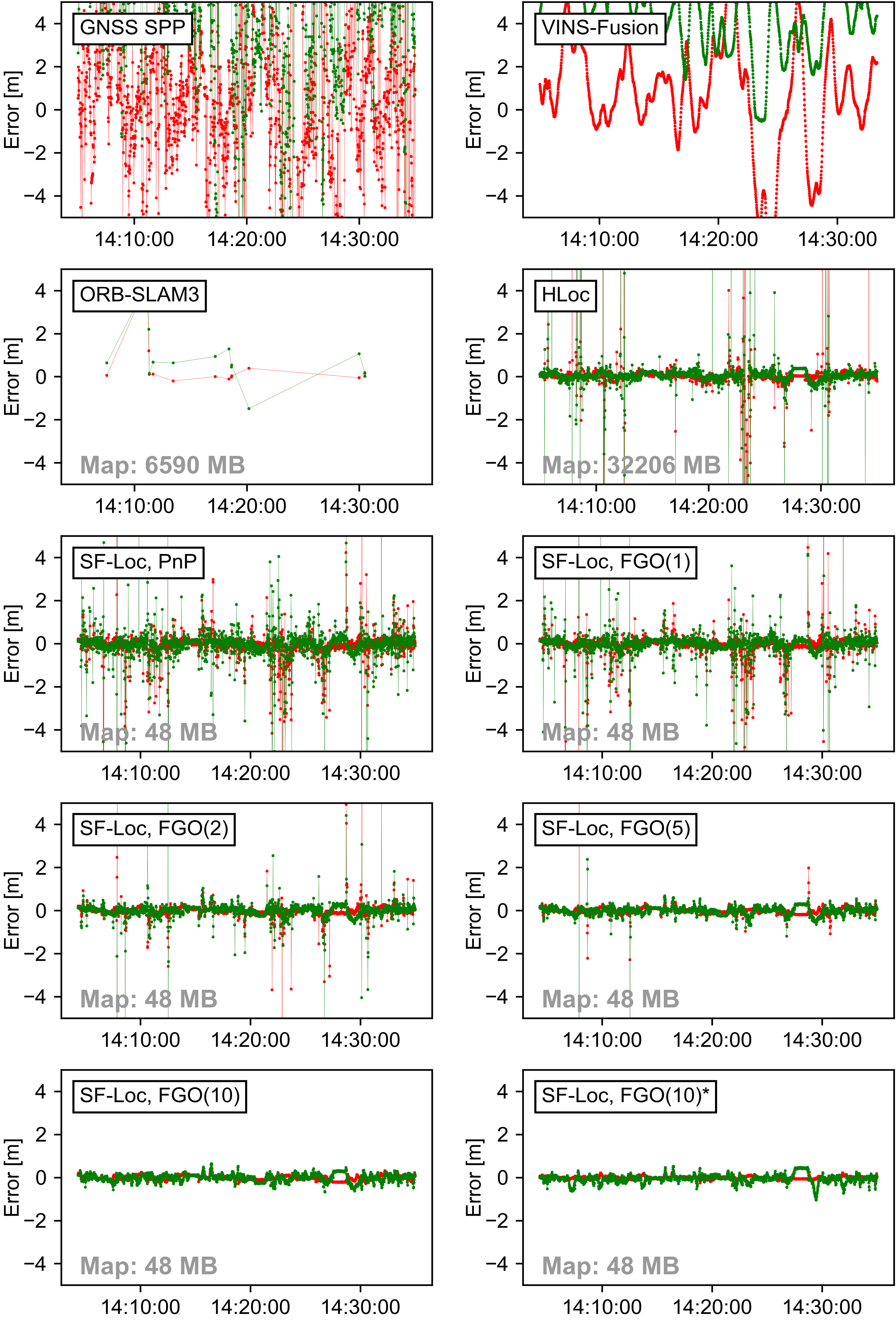}
	\caption{Horizontal position errors of the fine localization. 
	For SF-Loc, the $\Xi=0.4$ map is used and different multi-frame settings (1, 2, 5, 10) are employed. See TABLE \ref{table_fine} 
	for detailed description.}
	\label{fig_fine}
\end{figure}

\subsubsection{Fine Localization}
As to the evaluation of fine localization, we test the re-localization modes of ORB-SLAM3 and HLoc as the baselines, and also test GNSS-based schemes for reference, including single point positioning (SPP) and VINS-Fusion (stereo VIO + SPP).

 For ORB-SLAM3 and HLoc, the coarse-to-fine localization is performed based on their own pipeline and map data (corresponding to TABLE \ref{table_map}). For HLoc, EigenPlaces (50) is used for coarse localization and the SuperPoint + LightGlue scheme is used for fine localization.

For the proposed SF-Loc, the map data generated with different 
co-visibility thresholds ($\Xi=0.3/0.4$) are considered, and the coarse 
localization results of EigenPlaces50 (SAS10) are used. For fine localization, 
we test both single-frame and multi-frame  modes. For  
multi-frame fine localization, the matching results of $N$ sequential 
queries are used, and the user-side relative poses are provided by stereo VIO
(about 1\% relative translation error, consistent with 
typical performance\cite{ref_okvis}). Basically, the feature matching is performed using SuperPoint + LightGlue, and the factor graph optimization implementation in Sect. V-B is used. For comparison, we also test PnP with RANSAC upon 
the SF-Loc map. 


\renewcommand{\arraystretch}{0.9} 
\begin{table}[t]
	\centering
	\caption{Accuracy of the Fine Re-Localization. Horizontal Position Errors are Evaluated. $d_\text{VPR}$ Denotes the Median Recall Distance of Coarse Localization.}
	\label{table_fine}
	\setlength\tabcolsep{4pt} 
	\begin{threeparttable}
		\begin{tabular}{clcccc}
			\toprule
			\makecell[c]{Map\\\lbrack $d_\text{VPR}$\! (m)\rbrack } &
			\makecell[c]{Method\textsuperscript{1}} &
			\makecell[c]{Avail.\\@0.5 m} &
			\makecell[c]{Avail.\\@1.0 m} & 
			\makecell[c]{Avail.\\@5.0 m} & 
			\makecell[c]{RMSE\textsuperscript{2}\\(m)} \\ 
			\midrule
			- & SPP & 0.11\% & 0.44\% & 19.16\% & 8.059\\
			- & VINS-Fusion & 0.00\% & 0.00\% & 42.92\% & 5.628\\
			\midrule
			\makecell[c]{ORB \lbrack 3.12\rbrack} & \makecell[l]{ORB-SLAM3} & 0.44\% & 0.78\% & 1.00\% & 1.665\\
			\makecell[c]{SP \lbrack 2.42\rbrack} & \makecell[l]{HLoc} & 89.56\% & 93.23\% & 96.34\% & 0.505\\
			\midrule
			\multirow{6}{*}{\makecell[c]{SF ($\Xi$=0.4)\\ \lbrack 3.19\rbrack}} & PnP & 72.62\% & 86.53\% & 98.15\% & 0.839\\
		&	FGO (1) & 81.53\% & 90.11\% & 98.37\% & 0.722\\
		&	FGO (2) & 90.98\% & 96.20\% & 99.35\% & 0.452\\
		&	FGO (5) & 96.09\% & 99.40\% & 99.84\% & 0.259\\
		&	FGO (10) & \textbf{98.53\%} & \textbf{100.00\% }& \textbf{100.00\%} & \textbf{0.209}\\
		&	FGO (10)*  & 96.36\% & 99.89\% & 100.00\% & 0.219\\
			\midrule
			\multirow{6}{*}{\makecell[c]{SF ($\Xi$=0.3)\\ \lbrack 4.30\rbrack}}  & PnP & 65.73\% & 83.70\% & 97.45\% & 0.934\\
			& FGO (1) & 77.35\% & 87.29\% & 97.66\% & 0.830\\
		&	FGO (2) & 86.31\% & 93.86\% & 98.91\% & 0.614\\
		&	FGO (5) & 94.84\% & 98.26\% & 100.00\% & 0.306\\
		&	FGO (10) & \textbf{98.97\%} & \textbf{100.00\%} & \textbf{100.00\%} & \textbf{0.198}\\
		&	FGO (10)*  & 96.54\% & 99.84\% & 100.00\% & 0.227\\
			\bottomrule
		\end{tabular}
		\begin{tablenotes}
			\textsuperscript{1} For ORB-SLAM3, HLoc and SF-Loc without the ``*" notation, relative  pose estimation between the query frame and the map frame is evaluated. The ``*'' notation means that the absolute position error is evaluated, i.e., the pose error of the map frame (see Fig. \ref{fig_mapping}) is taken into account.
			
					\textsuperscript{2}  For RMSE calculation, only recall distance of coarse localization $<$20 m and final error $<$5 m cases are taken into account.
			
		\end{tablenotes}
	\end{threeparttable}
\end{table}

\renewcommand{\arraystretch}{0.9} 
\begin{table}[t]
	\centering
	\caption{
		Accuracy of Single-Frame Fine Localization using Different Matching Methods. 
		Using ``SF-Loc, FGO(1)'' Scheme ($\Xi=0.4$).}
	\label{table_mast}
	\setlength\tabcolsep{4pt} 
	\begin{threeparttable}
		\begin{tabular}{clcccc}
			\toprule
			\makecell[c]{Method\textsuperscript{1}} &
			\makecell[c]{Avail.\\@0.5 m} &
			\makecell[c]{Avail.\\@1.0 m} & 
			\makecell[c]{Avail.\\@5.0 m} & 
			\makecell[c]{RMSE\textsuperscript{1}\\(m)}& 
			\makecell[c]{Time Cost\\(ms)} \\ 
			\midrule
			SP + LG & \uwave{81.53\%} & 90.11\% & \underline{98.37\%} & \uwave{0.722} & 30 + 23\\
			AspanFormer & 	77.78\% & 87.83\% & 97.66\% & 0.882 & 125\\
			RoMa & \underline{87.02}\% & \underline{93.59}\% & \textbf{98.90}\% & \underline{0.719} & 371\\
			LoFTR & 	79.58\% & \uwave{90.49\%} & 	97.50\%& 	0.773 & 68\\
			MASt3R & 	\textbf{89.90\%} & \textbf{94.13\%} & \uwave{98.21\%}  & \textbf{0.580} & 307\\
			MASt3R\textsuperscript{\dag} & 32.54\% &52.36\% & 95.33\% &1.601 & 307\\
			\bottomrule
		\end{tabular}
		\begin{tablenotes}
			\textsuperscript{1}  For RMSE calculation, only $<$ 5 m cases are taken into account.\\
			\textsuperscript{\dag} Using predicted depth  instead of the depth in the map frame.
		\end{tablenotes}
	\end{threeparttable}
\end{table}

\begin{figure}[!t]
	\centering
	\includegraphics[width=7.8cm]{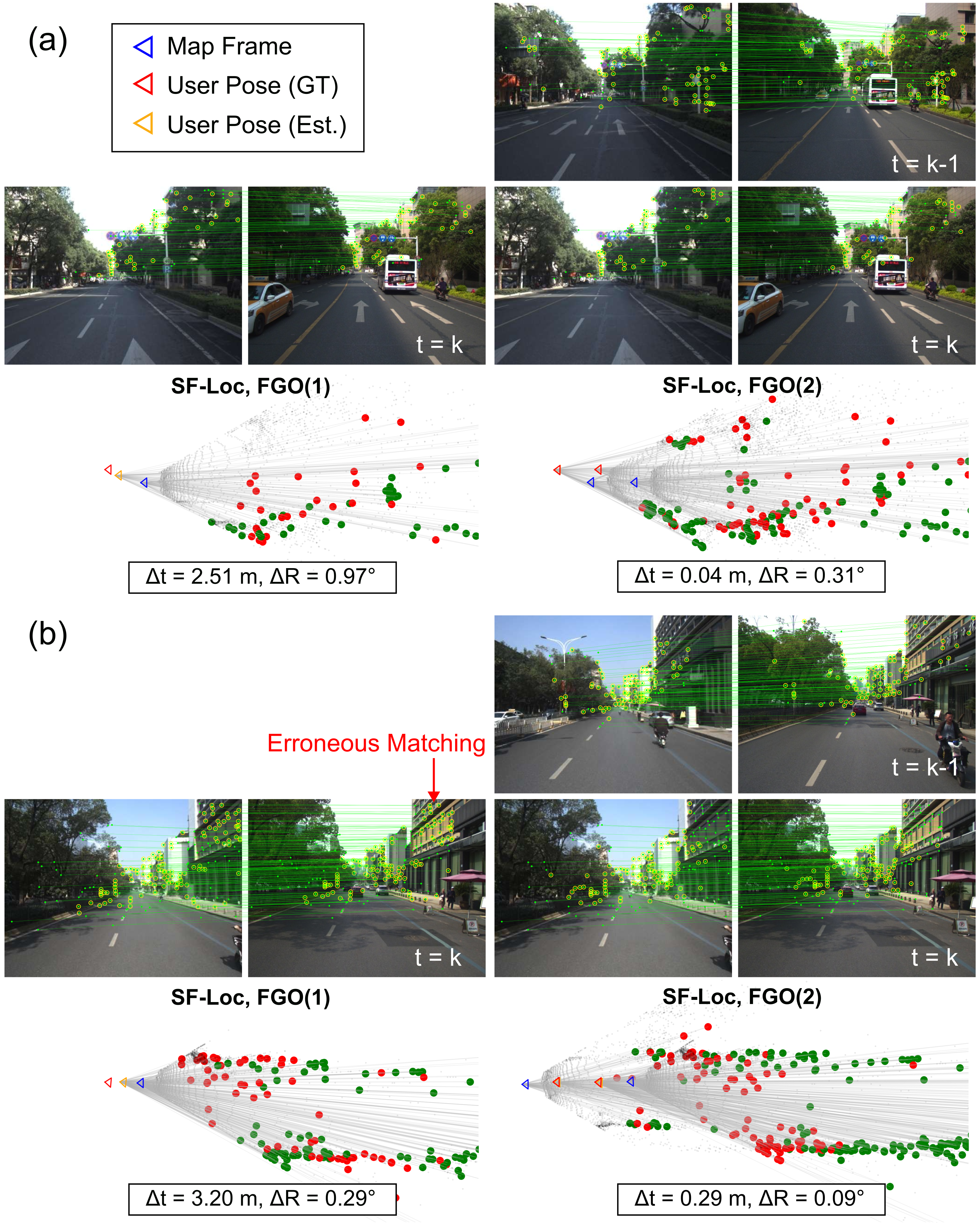}
	\caption{Examples of fine re-localization  based on single-/multi-frame observations, using ``SF-Loc, FGO(1)'' 
	and ``SF-Loc, FGO(2)'' schemes respectively.
	Position and attitude errors are provided in the attached boxes.}
	\label{fig_fine_cases}
\end{figure}

As can be seen from Fig. \ref{fig_fine} and TABLE \ref{table_fine}, 
ORB-SLAM3 can hardly perform successful re-localization, 
as the feature matching based on local intensity pattern  is greatly 
challenged by the significant appearance variation. 
For HLoc and SF-Loc, decimeter-level localization is 
achievable when the map matching is well performed, 
while outliers with several-meter error are also present, 
which indicates ambiguous data association. It is noted 
that HLoc performs better than single-frame SF-loc, which is reasonable 
as the HLoc map contains many more frames (see TABLE \ref{table_map}) 
and stores the fine descriptors extracted from  lossless raw images. 
In contrast, SF-Loc only uses highly compressed map frame 
information. 

 It is impressive that, the localization accuracy of 
 SF-Loc can be greatly improved by integrating multi-frame 
 information. The examples that multi-frame observations lead 
 to better data association and localization results are 
 depicted in Fig. \ref{fig_fine_cases}. As can be 
 seen, the dynamic objects and unstable features in 
 Fig. \ref{fig_fine_cases} (a), as well as the mismatches 
 caused by repetitive textures in Fig. \ref{fig_fine_cases} (b), lead
  to significant errors in single-frame localization.
 The introduction of multi-frame information  helps adjust 
 the inlier set of correspondences and improve the geometric configuration, 
 leading to more robust pose estimation. For the $\Xi=0.4$ map, SF-Loc with 10-frame FGO achieves 98.53\% availability@0.5 m 
 and an RMSE of 0.209 m. As to the absolute positioning accuracy considering the error of map frames, the availability@0.5 m is 96.36\% and the RMSE is 0.219 m, which mainly satisfy the demand of decimeter-level localization accuracy. When the map becomes sparser ($\Xi=0.3$), the multi-frame localization performance remains almost unchanged, which shows the possibility of harder map compression in the cost of some redundancy.
 
 In addition, based on the feature-free map with dense 
 depth, detector-free matching approaches can be 
 naturally applied in SF-Loc. 
 In TABLE \ref{table_mast}, we compare state-of-the-art detector-free 
 matching methods\cite{ref_loftr,ref_aspan,ref_roma,ref_mast3r} 
 with the default SuperPoint + LightGlue scheme.  The results demonstrate the possibility of achieving higher localization accuracy in cost of extra computational resources. We also consider a “map-free” localization strategy based on MASt3R’s metric depth prediction; However, its performance is relatively suboptimal, possibly due to domain discrepancies across datasets. This further underscores the value of incorporating metric-scale dense depth information into the visual structure frame map.

\begin{figure}[!t]
	\centering
	\includegraphics[width=7.4cm]{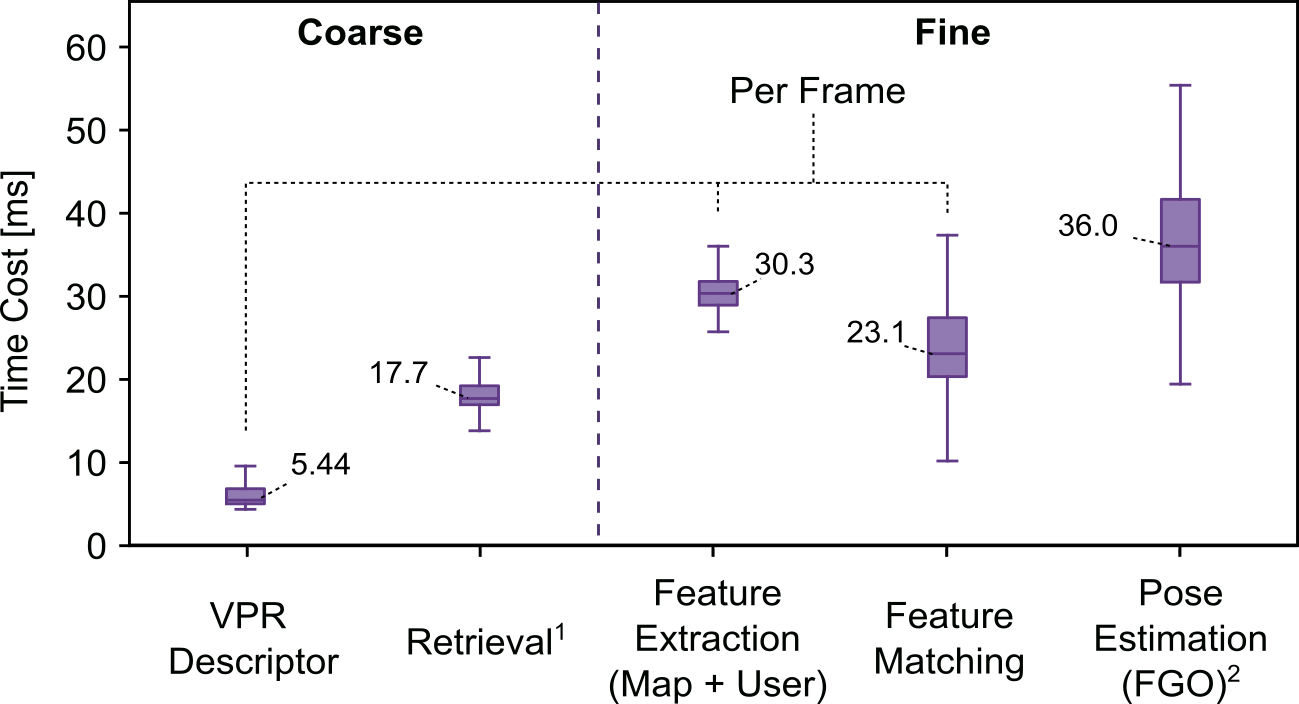}
	\caption{Time costs of different modules in 
	user-side localization of SF-Loc. It is noted 
	that  VPR descriptor computation, feature 
	extraction and matching are performed every 
	time a frame comes and stored, so it won't 
	lead to significant latency during multi-frame 
	processing. \textsuperscript{1} Using ``EigenPlaces50 (SAS10)'' scheme in TABLE \ref{table_coarse}.  
	\textsuperscript{2} Using ``SF-Loc, FGO(10)'' scheme in TABLE \ref{table_fine}.}
	\label{fig_time}
\end{figure}

Finally, we show the time consumption statistics of  user-side localization in Fig. 
\ref{fig_time}, which is tested on a laptop 
with Nvidia RTX 4080 Mobile GPU.
With an average processing time of around 110 ms per incoming image and an input frequency of 1 Hz,
the system demonstrates good real-time performance.

	\section{Conclusion}
	In this paper, we propose a vision-centered mapping and localization framework based on the map representation of visual structure frames. The highlight of the system is the lightweight map storage by making the frame data compact and keeping the sparsity, 
	while still being able to support high-recall, high-accuracy localization through the utilization of user-side multi-frame information.
	
	In our future work, we will extend the system to indoor scenarios, explore hybrid map representations, and further investigate the reliability issue of visual localization.
	

	\vspace{-1.5cm}
\begin{IEEEbiographynophoto}{Yuxuan Zhou} 
	received the M.Eng. degree in navigation engineering
	from Wuhan University, Wuhan, China, in 2022,
	where he is currently pursuing the Ph.D. degree with
	the School of Geomatics and Geodesy.
	His current research interests include integrated
	navigation systems and multi-sensor fusion.
\end{IEEEbiographynophoto}
\vspace{-1.25cm}
\begin{IEEEbiographynophoto}{Xingxing Li} received the Ph.D. degree from the Department
	of Geodesy and Remote Sensing, German Research
	Centre for Geosciences (GFZ).
	He is currently a Professor with Wuhan University.
	His current research interests include GNSS precise
	data processing and multi-sensor navigation.
\end{IEEEbiographynophoto}
\vspace{-1.25cm}
\begin{IEEEbiographynophoto}{Shengyu Li} 
	received the M.Eng. degree in navigation engineering
	from Wuhan University, Wuhan, China, in 2022,
	where he is currently pursuing the Ph.D. degree with
	the School of Geomatics and Geodesy.
	His current research interests include integrated
	navigation systems and multi-sensor fusion.
\end{IEEEbiographynophoto}
\vspace{-1.25cm}
\begin{IEEEbiographynophoto}{Chunxi Xia}received the M.Eng. degree in navigation
	engineering from Wuhan University, Wuhan, China,
	in 2023, where he is currently pursuing the Ph.D.
	degree with the School of Geomatics and Geodesy.
	His current research interests include vision-based
	navigation and multi-sensor fusion.
\end{IEEEbiographynophoto}
\vspace{-1.25cm}
\begin{IEEEbiographynophoto}{Xuanbin Wang} received the M.Eng. 
	degree in navigation engineering from 
	Wuhan University, Wuhan, China, in 2021. 
	He is currently pursuing the Ph.D. 
	degree with the School of Geomatics and 
	Geodesy, Wuhan University. His current 
	research focuses on integrated navigation 
	system and multi-sensor fusion. 
\end{IEEEbiographynophoto}
\vspace{-1.25cm}
\begin{IEEEbiographynophoto}{Shaoquan Feng}  received the M.Eng. degree in geodesy
	and survey engineering from Wuhan University,
	Wuhan, China, in 2020, where he is currently pursu
	ing the Ph.D. degree with the School of Geomatics
	and Geodesy.
	His current research interests include multi-object
	tracking, integrated navigation systems, and multi
	sensor fusion
\end{IEEEbiographynophoto}
	

\begin{thebibliography}{1}
\bibliographystyle{IEEEtran}
\bibitem{ref1} C. Cadena et al., ``Past, Present, and Future of Simultaneous Localization and Mapping: Toward the Robust-Perception Age,'' IEEE Trans. Robot., vol. 32, no. 6, pp. 1309-1332, Dec. 2016.
\bibitem{ref2} I. Ullah, D. Adhikari, H. Khan, M. S. Anwar, S. Ahmad, and X. Bai, ``Mobile robot localization: Current challenges and future prospective,'' Comput. Sci. Rev., vol. 53, p. 100651, 2024.
\bibitem{ref_giv1} X. Li et al., ``Continuous and precise positioning in urban environments by tightly coupled integration of GNSS, INS and vision,'' IEEE Robot. Autom. Lett., vol. 7, no. 4, pp. 11458–11465, 2022.
\bibitem{ref_gnss} W. Wen and L.-T. Hsu, ``Towards robust GNSS positioning and real-time kinematic using factor graph optimization,'' in ICRA 2021, pp. 5884–5890.
\bibitem{ref_map} A. Chalvatzaras, I. Pratikakis, and A. A. Amanatiadis, ``A survey on map-based localization techniques for autonomous vehicles,'' IEEE Trans. Intell. Veh., vol. 8, no. 2, pp. 1574–1596, 2022.
\bibitem{ref_pr}P. Yin et al., ``General Place Recognition Survey: Towards Real-World Autonomy,'' May 08, 2024, arXiv: arXiv:2405.04812.
\bibitem{ref_bevlocator} Z. Zhang et al., ``BEV-Locator: An End-to-end Visual Semantic Localization Network Using Multi-View Images,'' Nov. 27, 2022, arXiv: arXiv:2211.14927.
\bibitem{ref_orb} C. Campo et al., ``ORB-SLAM3: An Accurate Open-Source Library for Visual, Visual-Inertial, and Multimap SLAM,'' IEEE Trans. Robot., vol. 37, no. 6, pp. 1874-1890, Dec. 2021.
\bibitem{ref_hloc} P. E. Sarlin et al., ``From coarse to fine: Robust hierarchical localization at large scale.'' in CVPR 2019, pp. 12716-12725.
\bibitem{ref_grid}  L. Xu, C. Feng, V. R. Kamat, and C. C. Menassa, ``An Occupancy Grid Mapping enhanced visual SLAM for real-time locating applications in indoor GPS-denied environments,'' Autom. Constr., vol. 104, pp. 230–245, Aug. 2019.
\bibitem{ref_locnerf} D. Maggio et al., ``Loc-NeRF: 
Monte Carlo Localization using Neural Radiance Fields,'' arXiv preprint arXiv:2209.09050, 2022.
\bibitem{ref_splatloc} H. Zhai et al., ``SplatLoc: 3D Gaussian Splatting-based Visual Localization for Augmented Reality,'' 
Sep. 21, 2024, arXiv: arXiv:2409.14067.
\bibitem{ref_vins} T. Qin, P. Li, and S. Shen, ``Vins-mono: A robust and versatile monocular visual-inertial state estimator,'' IEEE Trans. Robot., vol. 34, no. 4, pp. 1004-1020, 2018.
\bibitem{ref_okvis} S. Leutenegger et al., ``Keyframe-based visual-inertial odometry using nonlinear optimization,''IJRR, vol. 34, no. 3, pp. 314-334, 2014.
\bibitem{ref_vow} K. Wu, C. Guo, G. Georgiou, and S. I. Roumeliotis. ``Vins on wheels,'' in ICRA 2017, pp. 5155-5162.
\bibitem{ref_maplab2} A. Cramariuc et al., ``maplab 2.0–a modular and multi-modal mapping framework,'' IEEE Robot. Autom. Lett., vol. 8, no. 2, pp. 520–527, 2022.
\bibitem{ref_kimeram} Y. Tian et al., ``Kimera-multi: Robust, distributed, dense metric-semantic slam for multi-robot systems,'' IEEE Trans. Robot., vol. 38, no. 4, 2022.
\bibitem{ref_nicer} Z. Zhu et al., ``Nicer-slam: Neural implicit scene encoding for rgb slam,'' Feb. 2023, arXiv:2302.03594.
\bibitem{ref_gsslam} H. Matsuki, R. Murai, P. H. Kelly, and A. J. Davison, ``Gaussian splatting slam,'' in CVPR 2024, pp. 18039–18048.
\bibitem{ref_dpvo}  Z. Teed, L. Lipson, and J. Deng, ``Deep patch visual odometry,'' in NeurIPS 2024.
\bibitem{ref_droid} Z. Teed and J. Deng. ``Droid-slam: Deep visual slam for monocular, stereo, and rgb-d cameras,'' in NeurIPS 2021, pp. 16558-16569.
\bibitem{ref_ssfearture} S. Li et al., ``Quantized Self-Supervised Local Feature for Real-Time Robot Indirect VSLAM,'' in IEEE/ASME TMech, vol. 27, no. 3, pp. 1414-1424, June 2022.
\bibitem{ref_e2e_vo1} S. Wang, R. Clark, H. Wen, and N. Trigoni, ``Deepvo: Towards end-to-end visual odometry with deep recurrent convolutional neural networks,'' in ICRA 2017, pp. 2043-2050.
\bibitem{ref_bow} D. Gálvez-López, J. D. Tardos. 
``Bags of binary words for fast place recognition 
in image sequences,'' IEEE Trans. Robot., vol. 28, no.5, pp. 1188-1197, 2012.
\bibitem{ref_vlad} H. Jégou et al., ``Aggregating local descriptors into a compact image representation,'' in CVPR 2010, pp. 3304–3311.
\bibitem{ref_seqslam} M. J. Milford and G. F. Wyeth, ``SeqSLAM: Visual route-based navigation for sunny summer days and stormy winter nights,'' in ICRA 2012, pp. 1643–1649.
\bibitem{ref_netvlad} R. Arandjelović et al., ``NetVLAD: CNN Architecture for Weakly Supervised Place Recognition,'' IEEE TPAMI, vol. 40, no. 6, pp. 1437-1451, June 2018.
\bibitem{ref_transVPR} R. Wang et al., ``Transvpr: Transformer-based place recognition with multi-level attention aggregation,'' in CVPR 2022, pp. 13648–13657.
\bibitem{ref_cosplace} G. Berton, C. Masone, and B. Caputo, ``Rethinking visual geo-localization for large-scale applications,'' in CVPR 2022, pp. 4878–4888.
\bibitem{ref_eigenplaces} G. Berton et al., ``Eigenplaces: Training viewpoint robust models for visual place recognition,'' in ICCV 2023, pp. 11080–11090.
\bibitem{ref_anyloc} N. Keetha et al., ``AnyLoc: Towards Universal Visual Place Recognition,'' IEEE Robot. Autom. Lett., vol. 9, no. 2, pp. 1286-1293, Feb. 2024.
\bibitem{ref_vins_fusion} T. Qin et al., ``A general optimization-based framework for global pose estimation with multiple sensors,'' arXiv:1901.03642, 2019.    
\bibitem{ref_mapfree} E. Arnold et al., ``Map-Free Visual Relocalization: Metric Pose Relative to a Single Image,'' in ECCV 2022.
\bibitem{ref_obj1} 	 S. Yang and S. Scherer, ``Cubeslam: Monocular 3-d object slam,'' IEEE Trans. Robot., vol. 35, no. 4, pp. 925-938, 2019.
\bibitem{ref_obj4} Y. Yu et al., ``Accurate and Robust Visual Localization System in Large-Scale Appearance-Changing Environments,'' in IEEE/ASME TMech, vol. 27, no. 6, pp. 5222-5232, Dec. 2022.
\bibitem{ref_onet} P.-E. Sarlin et al., ``Orienternet: Visual localization in 2d public maps with neural matching,'' in CVPR 2023, pp. 21632–21642.
\bibitem{ref_3dgsreloc} P. Jiang, G. Pandey, and S. Saripalli, ``3DGS-ReLoc: 
3D Gaussian Splatting for Map Representation and Visual ReLocalization,'' arXiv preprint arXiv:2403.11367, 2024.
\bibitem{ref_jpeg} Wallace, Gregory K. "The JPEG still picture compression standard." Communications of the ACM, vol. 34, no. 4, pp. 30-44, 1991.
\bibitem{ref_gtsam} F. Dellaert and M. Kaess, Factor Graphs for Robot Perception. Foundations, 2017.
\bibitem{ref_dbaf}  Y. Zhou et al., ``DBA-Fusion: Tightly Integrating Deep Dense Visual Bundle Adjustment With Multiple Sensors for Large-Scale Localization and Mapping,''  IEEE RAL, vol. 9, no. 7, pp. 6138–6145, Jul. 2024.
\bibitem{ref_imu} C. Forster, L. Carlone, F. Dellaert and D. Scaramuzza, ``On-manifold preintegration for real-time visual–inertial odometry,'' IEEE Trans. Robot., vol. 33, no. 1, pp. 1-21, Feb. 2017.
\bibitem{ref_sp} D. DeTone, T. Malisiewicz, and A. Rabinovich, ``Superpoint: Self-supervised interest point detection and description,'' in CVPRW 2018, pp. 224-236.
\bibitem{ref_lg} P. Lindenberger, P.-E. Sarlin, and M. Pollefeys, ``Lightglue: Local feature matching at light speed,'' in ICCV 2023, pp. 17627–17638.
\bibitem{ref_sg} P.-E. Sarlin, D. DeTone, T. Malisiewicz, and A. Rabinovich, ``Superglue: Learning feature matching with graph neural networks,'' in CVPR 2020, pp. 4938–4947. 
\bibitem{ref_mixvpr} A. Ali-Bey, B. Chaib-Draa, and P. Giguere, ``Mixvpr: Feature mixing for visual place recognition,'' in WACV 2023, pp. 2998–3007.
\bibitem{ref_colmap} J. L. Schonberger and J.-M. Frahm, ``Structure-from-motion revisited,'' in CVPR 2016, pp. 4104–4113. 
\bibitem{ref_resnet} K. He, X. Zhang, S. Ren, and J. Sun, ``Deep residual learning for image recognition,'' in CVPR 2016, pp. 770–778.
\bibitem{ref_loftr} J. Sun et al., ``LoFTR: Detector-free local feature matching with transformers,'' in CVPR 2021, pp. 8922–8931.
\bibitem{ref_aspan} H. Chen et al., ``ASpanFormer: Detector-Free Image Matching with Adaptive Span Transformer,'' in 
ECCV 2022, pp. 20–36.
\bibitem{ref_roma} J. Edstedt et al., ``RoMa: Robust Dense Feature Matching,'' in CVPR 2024, pp. 19790–19800.
\bibitem{ref_mast3r} V. Leroy, Y. Cabon, and J. Revaud, ``Grounding Image Matching in 3D with MASt3R,'' in ECCV 2024, pp. 71–91.



	\end{thebibliography}
\end{document}